\pdfoutput=1
\documentclass[11pt]{article}

\usepackage[]{acl}

\usepackage{times}
\usepackage{latexsym}

\usepackage{paralist}
\usepackage[T1]{fontenc}
\usepackage[utf8]{inputenc}

\usepackage{microtype}

\usepackage{inconsolata}

\usepackage{amsmath,amssymb,amsfonts}
\usepackage{url,enumitem}
\usepackage{booktabs}
\usepackage{xspace}
\usepackage{graphicx}
\usepackage{subcaption}
\usepackage{comment}
\usepackage[normalem]{ulem}
\usepackage{mathtools}
\usepackage{svg}
\usepackage{algorithm,algorithmicx,algpseudocode}

\usepackage{listings}
\newcommand{\Sref}[1]{\S\ref{#1}}

\DeclareMathOperator{\argmax}{argmax}

\newcommand{\methodname}[1]{\textsc{Ssd-LM}}

\let\orgautoref\autoref

\renewcommand{\autoref}[1]{\def\equationautorefname{Eq.}\orgautoref{#1}}

\title{
\methodname{}: Semi-autoregressive Simplex-based Diffusion Language Model for Text Generation and Modular Control
}

\author{
  Xiaochuang Han$^\spadesuit$ \quad \quad \quad \quad Sachin Kumar$^\clubsuit$ \quad \quad \quad \quad Yulia Tsvetkov$^\spadesuit$\\
  $^\spadesuit$Paul G.~Allen School of Computer Science \& Engineering, University of Washington \\
  $^\clubsuit$Language Technologies Institute, Carnegie Mellon University \\
        {\tt \{xhan77, yuliats\}@cs.washington.edu$^\spadesuit$ \quad sachink@cs.cmu.edu$^\clubsuit$}
}

\begin{document}
\maketitle

\begin{abstract}

Despite the growing success of diffusion models in continuous-valued domains (e.g., images), similar efforts for discrete domains such as text have yet to match the performance of autoregressive language models. In this work, we present \methodname{}---a diffusion-based language model with two key design choices. First, \methodname{} is \emph{semi-autoregressive}, iteratively generating blocks of text, allowing for flexible output length at decoding time while enabling local bidirectional context updates. Second, it is \emph{simplex-based}, performing diffusion on the natural vocabulary space rather than a learned latent space, allowing us to incorporate classifier guidance and modular control using off-the-shelf classifiers without any adaptation. We evaluate \methodname{} on unconstrained text generation benchmarks, and show that it matches or outperforms strong autoregressive GPT-2 models across standard quality and diversity metrics, while vastly outperforming diffusion-based baselines. On controlled text generation, \methodname{} also outperforms competitive baselines, with an extra advantage in modularity.\footnote{
Our code and models can be found at \url{https://github.com/xhan77/ssd-lm}. 
% \url{ANONYMIZED}
}

\end{abstract}

% PLAIN TEXT ABSTRACT, final camera-ready
% Despite the growing success of diffusion models in continuous-valued domains (e.g., images), similar efforts for discrete domains such as text have yet to match the performance of autoregressive language models. In this work, we present SSD-LM---a diffusion-based language model with two key design choices. First, SSD-LM is semi-autoregressive, iteratively generating blocks of text, allowing for flexible output length at decoding time while enabling local bidirectional context updates. Second, it is simplex-based, performing diffusion on the natural vocabulary space rather than a learned latent space, allowing us to incorporate classifier guidance and modular control using off-the-shelf classifiers without any adaptation. We evaluate SSD-LM on unconstrained text generation benchmarks, and show that it matches or outperforms strong autoregressive GPT-2 models across standard quality and diversity metrics, while vastly outperforming diffusion-based baselines. On controlled text generation, SSD-LM also outperforms competitive baselines, with an extra advantage in modularity.

\section{Introduction}
Diffusion models \citep{sohl2015deep}, trained to iteratively refine noised inputs, have recently emerged as powerful tools for generative modeling in several continuous-valued domains such as images \citep{ho2020denoising}, audio \citep{Kong2021DiffWaveAV}, video \citep{Ho2022VideoDM}, among others. Attempts to adapt them for discrete domains such as text data, however, have only had limited success: prior work have shown to be promising on specialized cases and small datasets \citep{hoogeboom2021argmax,Austin2021StructuredDD,Li2022DiffusionLMIC,Chen2022AnalogBG}, 
% with byte or character based approaches [cite] or controllability on small scale datasets and tasks [cite], 
but diffusion models for text still underperform (and thus are not widely adopted) compared to autoregressive language models (AR-LMs) which remain the state-of-the-art general purpose text generators \citep{Radford2019LanguageMA,Brown2020LanguageMA}.  

% Towards understanding this performance gap in the existing literature, 
Despite potential advantages of diffusion models for text, there are two key challenges. %barriers in their adoption. 
First, diffusion models generate text non-autoregressively, i.e., they generate (and update) the entire sequence simultaneously rather than token by token left-to-right. Although this property 
% non-autoregressive generation 
is useful in practice since each output token is informed by a broader bi-directional %(latent) 
context  \citep{Lee2018DeterministicNN,Ghazvininejad2019MaskPredictPD}, it requires pre-defining an output sequence length. This limits the flexibility and applicability of trained models. 
%, since a model trained with short sequence lengths is not usable for generating longer sequences. 
% On the other hand, non-autoregressive training with long sequences is difficult to optimize, expensive, and less scalable to large datasets. 
On the other hand, non-autoregressive training with long sequences is expensive and difficult to optimize. 
In this work, we propose a \emph{semi-autoregressive} solution which strikes a balance between length flexibility and the ability to alter previously generated tokens.

A major advantage of diffusion models over the current standard of autoregressive LMs is their post-hoc controllability using guidance from auxiliary models such as style classifiers \citep{Dhariwal2021DiffusionMB}. However, controllability is hard to achieve without compromises in modularity in diffusion-based LMs for text. To enable diffusion generation into discrete text rather than continuous modalities, prior approaches have employed different approximations, e.g., training with embeddings, character, or byte-level methods \citep{Li2022DiffusionLMIC,hoogeboom2021argmax,Austin2021StructuredDD,Chen2022AnalogBG}. In contrast, existing mainstream LMs and the guidance classifiers they derive often operate at a sub-word level with  sub-word representations trained jointly with the language model \citep{Devlin2019BERTPO,Liu2019RoBERTaAR,Raffel2020ExploringTL}. Subsequently, changing the input representations to characters or embeddings requires developing guidance models from scratch, which can be expensive or infeasible in many cases. In this work, we propose a \emph{simplex-based} solution which enables the diffusion over discrete texts while maintaining the advantages of diffusion models with plug-and-control  guidance models.

In sum, to enable diffusion-based LMs for text we present \methodname{} (\Sref{sec:method}), addressing the above two challenges. \methodname{} is trained to generate text semi-autoregressively---generating blocks of tokens left-to-right with bidirectional context within the block---which offers the benefits of both AR-LMs and diffusion models. It supports training with and generating variable-length sequences. At the same time, it allows refinement within the token block, in contrast to token-level autoregressive decoding where previously generated tokens cannot be modified at all. 
\methodname{} uses the same tokenization as popular AR-LMs, representing discrete text via a distribution (or simplex) defined over the vocabulary and is trained to reconstruct texts from noisy versions of the distributions. 
Due to its underlying representation, our method also offers an easy and modular way of guided (controlled) generation using off-the-shelf text classifiers under the minimal assumption of shared tokenizer.

Our evaluation experiments show, for the first time, 
% to our knowledge, 
that a diffusion-based LM matches or outperforms strong AR-LMs on standard text generation benchmarks (\Sref{sec:experiments}). 
We evaluate \methodname{} on two tasks: (1) unconstrained prompt-based generation substantially outperforming existing diffusion LM approaches and performing on par with or outperforming strong autoregressive LM GPT-2~\citep{Radford2019LanguageMA} %for the first time that a diffusion LM can outperform strong autoregressive baselines (GPT-2) 
on both quality and diversity (\Sref{sec:experiments_natural_gen}); and (2) controlled text generation with guidance from off-the-shelf classifiers (no post-hoc training/adaptation) outperforming competitive controlled text generation baselines  (\Sref{sec:experiments_controlled_gen}). 
% We release a demo of \methodname{} based on HuggingFace's code base to facilitate simple adoption and extension: {\small \url{https://github.com/xhan77/ssd-lm}}.

\section{Background}

\subsection{Diffusion model}
Since their inception as image generators, diffusion models (and their cousins score-based models \citep{song2019generative}) have been widely adopted as high-quality generative models for multiple data modalities. Here, we briefly describe a simplified view of a canonical method, denoising diffusion probabilistic models \citep[DDPM]{ho2020denoising} which we adapt in this work for text generation. We assume a given dataset $\mathcal{D}=\{{}^{1}\boldsymbol{x}_0, \ldots, {}^{N}\boldsymbol{x}_0\}$ of continuous valued items ${}^{i}\boldsymbol{x}_0$ (e.g., pixel values of an image) henceforth referred to as $\boldsymbol{x}_0$ for simplicity.

\paragraph{Training}
Training a diffusion model first involves adding a series of Gaussian noise to the original data $\boldsymbol{x}_0$, through $T$ timesteps:
\begin{align}
    % p(\boldsymbol{x}_t \mid \boldsymbol{x}_0) &= \mathcal{N}(\boldsymbol{x}_t; \sqrt{\Bar{\alpha}_t} \boldsymbol{x}_0, (1-\Bar{\alpha}_t) \textbf{I})
    \boldsymbol{x}_t &= \sqrt{\Bar{\alpha}_t} \boldsymbol{x}_0 + \sqrt{1-\Bar{\alpha}_t} \boldsymbol{\epsilon}_t \label{eq:1}
\end{align}
where $t \in (1, T)$ and $\boldsymbol{\epsilon}_t \sim \mathcal{N}(\boldsymbol{0}, \mathbf{I})$.  
% $\Bar{\alpha}_t$ depends on a schedule of $\alpha_{t}$, where 
$\Bar{\alpha}_t = \prod_{t'=1}^{t} \alpha_{t'}$, where $\alpha_{t'}$ follow a predefined schedule such that $\Bar{\alpha}_t \to 0$ as $t \to T$. This process is called \emph{forward diffusion}.
% $f_{\theta}(\boldsymbol{x}_t, t)$
A diffusion model (parameterized by $\theta$) is trained to reverse this forward process by predicting the added noise $\boldsymbol{\epsilon}_t$ given $\boldsymbol{x}_t$ with the following loss:
\begin{align}
    \mathcal{L}(\theta) = \mathbb{E}_{t \sim \mathcal{U}(1, T)} \lVert \epsilon_{\theta}(\boldsymbol{x}_t, t) - \boldsymbol{\epsilon}_t \rVert^2 \label{eq:2}
\end{align}

\paragraph{Inference}
To get an output from this model, we sample  $\boldsymbol{x}_T \sim \mathcal{N}(\boldsymbol{0}, \mathbf{I})$ and iteratively reconstruct a sample $\boldsymbol{x}_0$ by going back in time,
\begin{align}
    % p(\boldsymbol{x}_t \mid \boldsymbol{x}_0) &= \mathcal{N}(\boldsymbol{x}_t; \sqrt{\Bar{\alpha}_t} \boldsymbol{x}_0, (1-\Bar{\alpha}_t) \textbf{I})
    \boldsymbol{x}_{t-1} &= \frac{1}{\sqrt{\alpha_t}} (\boldsymbol{x}_{t} - \frac{1-\alpha_t}{\sqrt{1-\Bar{\alpha}_t}} \epsilon_{\theta}(\boldsymbol{x}_t, t)) \label{eq:3}
\end{align}
for $t = T, \ldots, 1$.\footnote{We omit an additional noise term $z$ here for simplicity, which is present in DDPM but not in another variant DDIM \cite{Song2021DenoisingDI}.} 
The key obstacle in using vanilla diffusion models directly as text generators is that language consists of discrete tokens, i.e., a non-continuous   $\boldsymbol{x}_{0}$ to which a continuous valued Gaussian noise cannot be added. We propose a straightforward and effective solution by treating tokens as continuous valued simplexes over the vocabulary~\citep{hoang-etal-2017-towards}. 
Other existing methods addressing this problem are discussed in \Sref{sec:related_work}.

\subsection{Autoregressive LM}
An autoregressive LM 
% like GPT-2 \citep{Radford2019LanguageMA}
model optimizes for the likelihood of a sequence of tokens $w^0, \ldots, w^{L-1}$.
\begin{align}
    p_\theta(\boldsymbol{w}^{0:L}) = \prod_{c=0}^{L-1} p_\theta(w^{c} \mid \boldsymbol{w}^{<c}) \label{eq:4}
\end{align}
To decode from AR-LMs, one can provide a context $\boldsymbol{w}^{<c}$ and decode the next token $w^{c}$ iteratively by predicting $p_\theta(w^{c} \mid \boldsymbol{w}^{<c})$ and sampling from it to get the discrete token \citep{fan2018hierarchical,holtzman2019curious}. % and then increasing $c$.
Prior work has shown that these decoding approaches (and by extension the LMs themselves) are prone to degrade when generating long sequences and often devolve into repeating subsequences~\citep{holtzman2019curious,Meister2022LocallyTS}. In addition, such LMs do not provide a natural way to incorporate sequence-level control as tokens are generated one at a time without the ability to modify previously generated tokens~\citep{dathathri2019plug,Kumar2022ConstrainedSF}.
% Issues of autoregressive decoding token-by-token include repetitions in long generations \citep{holtzman2019curious,Meister2022LocallyTS} and difficulty of controlling generations by target attributes \citep{dathathri2019plug,Kumar2022ConstrainedSF}. 
In this work, we present a method to train a semi-autoregressive LM that decodes blocks of $B$ tokens at a time, 
% modeling $p_\theta(\boldsymbol{w}^{c:c+B} \mid \boldsymbol{w}^{<c})$.
alleviating said issues with the support of diffusion models. Existing literature addressing the two issues individually are discussed in \Sref{sec:related_work}.

\section{\methodname{}~~\includegraphics[scale=0.11]{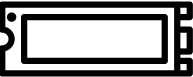}}
\label{sec:method}

\begin{figure}[t]
    \centering
    \includegraphics[width=0.44\textwidth]{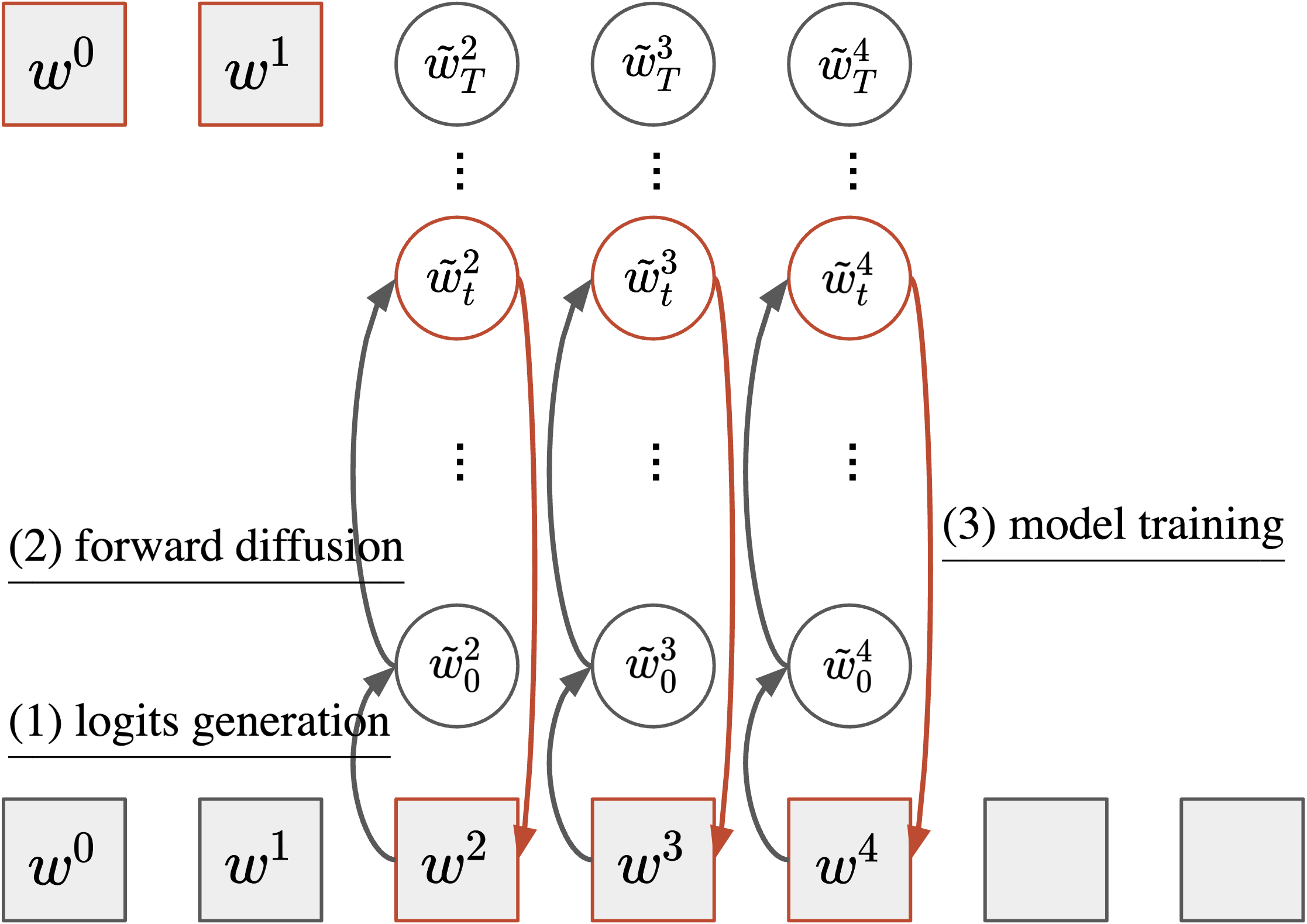}
    \caption{Training \methodname{} (a snapshot at context size $c=2$, block size $B=3$). Horizontal axis represents the order of tokens. Vertical axis represents the diffusion timesteps. Shade means observable variables. Square means discrete vocabulary, while circle means continuous logits. Red components are inputs to the learning model $\theta$.
    }
    \label{fig:training_fig}
    \vspace{-0.6em}
\end{figure}

\begin{figure}[t]
    \centering
    \includegraphics[width=0.44\textwidth]{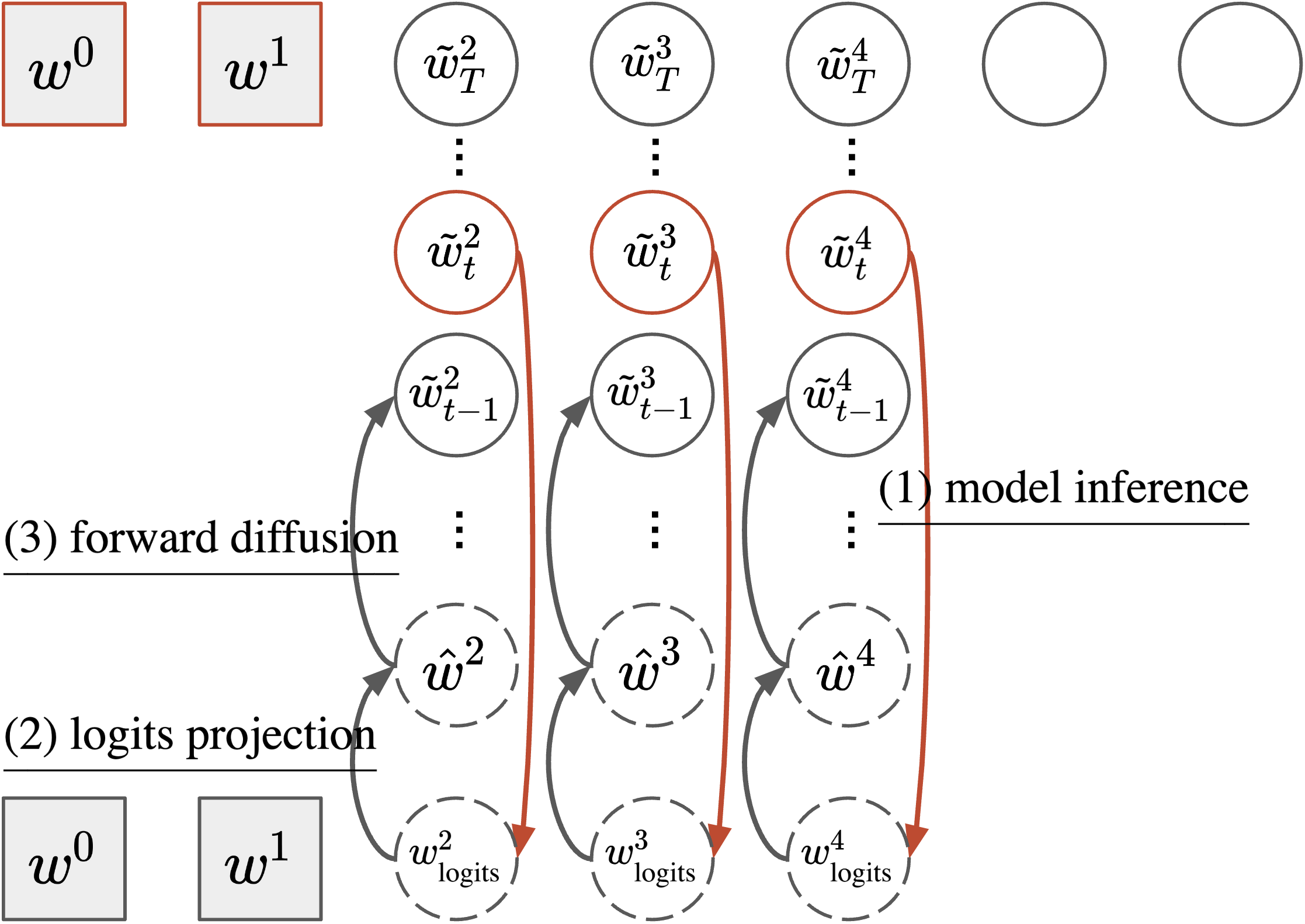}
    \caption{Decoding from \methodname{} (continuing \autoref{fig:training_fig}). 
    % Horizontal axis represents the order of tokens. Vertical axis represents the diffusion timesteps. Shade means observable variables. Square means discrete vocabulary. Circle means distribution of vocabulary. 
    Red components are inputs to the learned model $\theta$. Dash means intermediate variables.
    }
    \label{fig:decoding_fig}
    \vspace{-0.8em}
\end{figure}

We introduce \methodname{}---\textbf{S}emi-autoregressive \textbf{S}implex-based \textbf{D}iffusion Language Model--- adapting key components from both autoregressive LM and vanilla diffusion models. Conceptually, \methodname{} uses diffusion model to decode $\boldsymbol{w}^{c:c+B}$, a block  of tokens of length $B$, given a Gaussian noise and a context $\boldsymbol{w}^{<c}$ of length $c$. 
% That is, unlike an AR-LM that models $p_\theta(w^{c} \mid \boldsymbol{w}^{<c})$, \methodname{} models $p_\theta(\boldsymbol{w}^{c:c+B} \mid \boldsymbol{w}^{<c})$. 
% At decoding time, it semi-autoregressively generates blocks from left to right iteratively as $c$ increases.
We show an intuitive diagram and pseudo-code for the training and decoding algorithm of \methodname{} in \autoref{fig:training_fig}, \autoref{fig:decoding_fig}, and \autoref{fig:algo}.

\algrenewcommand\algorithmicindent{0.5em}%
\begin{figure*}[t]
\begin{minipage}[t]{0.495\textwidth}
\begin{algorithm}[H]
  \caption{Training} \label{alg:training}
  \small
  \begin{algorithmic}[1]
    \Repeat
      \State $\boldsymbol{w}^{0:L} \sim q(\boldsymbol{w}^{0:L})$
      \State $c \sim \mathrm{Uniform}(\{1, \dotsc, L-B\})$
      \State $\Tilde{\boldsymbol{w}}_0^{c:c+B} = \operatorname{logits-generation}(\boldsymbol{w}^{c:c+B})$
      \State $t \sim \mathrm{Uniform}(\{1, \dotsc, T\})$
      \State $\boldsymbol{\epsilon} \sim \mathcal{N}(\boldsymbol{0}, K^2\mathbf{I})$
      \State $\Tilde{\boldsymbol{w}}_t^{c:c+B} = \sqrt{\Bar{\alpha}_t} \Tilde{\boldsymbol{w}}_0^{c:c+B} + \sqrt{1-\Bar{\alpha}_t} \boldsymbol{\epsilon}$
      \State Take gradient descent step on
      \Statex $\qquad \nabla_\theta [-\sum_{j=c}^{c+B-1}\log p_\theta(w^{j} \mid \Tilde{\boldsymbol{w}}_t^{c:c+B}, \boldsymbol{w}^{<c})]$
    \Until{converged}
  \end{algorithmic}
\end{algorithm}
\end{minipage}
\hfill
\begin{minipage}[t]{0.495\textwidth}
\begin{algorithm}[H]
  \caption{Decoding (at a given $c$)} \label{alg:sampling}
  \small
  \begin{algorithmic}[1]
    \vspace{.04in}
    \State $\Tilde{\boldsymbol{w}}_T^{c:c+B} \sim \mathcal{N}(\boldsymbol{0}, K^2\mathbf{I})$
    \For{$t=T, \dotsc, 1$}
    \State $\boldsymbol{w}_{\text{logits}}^{c:c+B} = \operatorname{logits}_\theta(\boldsymbol{w}^{c:c+B} \mid \Tilde{\boldsymbol{w}}_t^{c:c+B}, \boldsymbol{w}^{<c})$
    \State $\hat{\boldsymbol{w}}^{c:c+B} = \operatorname{logits-projection}(\boldsymbol{w}_{\text{logits}}^{c:c+B})$ if uncontrolled, else $\hat{\boldsymbol{w}}^{c:c+B} = \operatorname{logits-projection}(\boldsymbol{w}_{\text{logits}}^{c:c+B} + \lambda \nabla_{\boldsymbol{w}} f_{\phi}(\cdot))$
    \State $\boldsymbol{z} \sim \mathcal{N}(\boldsymbol{0}, K^2\mathbf{I})$
    \State $\Tilde{\boldsymbol{w}}_{t-1}^{c:c+B} = \sqrt{\Bar{\alpha}_{t-1}} \hat{\boldsymbol{w}}^{c:c+B} + \sqrt{1-\Bar{\alpha}_{t-1}} \boldsymbol{z}$
    \EndFor
    \State \textbf{return} $\argmax \Tilde{\boldsymbol{w}}_{0}^{c:c+B}$
    \vspace{.006in}
  \end{algorithmic}
\end{algorithm}
\end{minipage}

\caption{Training and decoding algorithms for \methodname{}. The training algorithm starts with sampling a sequence from the pretraining data $q(\boldsymbol{w}^{0:L})$. The decoding algorithm can be applied $m$ iterations to obtain a $m \cdot B$-token generation, with the returned $B$ tokens at each iteration appended to the previous generation, increasing $c$.}
\label{fig:algo}
\vspace{-0.8em}
\end{figure*}

\subsection{Training}
\label{sec:method_training}
\paragraph{Continuous data representation}
To build a continuous representation for discrete tokens, we adopt an \emph{almost-one-hot} simplex representation over the model's vocabulary $V$. We define a simple operation $\operatorname{logits-generation}(.)$ to map a token $w$ to $\Tilde{\boldsymbol{w}} \in \{ -K, +K \}^{|V|}$ as follows.
\vspace{-0.5em}
\begin{align}
\Tilde{w}_{(i)} = \begin{cases}
+K \text{ when } w = V_{(i)}\\
-K \text{ when } w \neq V_{(i)}
\end{cases} \label{eq:5}
\end{align}
where $i$ is the index of the vocabulary. We call $\Tilde{\boldsymbol{w}}$ the logits for token $w$, and $\operatorname{softmax}(\Tilde{\boldsymbol{w}})$ gives a probability simplex over the vocabulary $V$, with a probability mass concentrated on the token $w$. There is no learnable parameter in this mapping.

\paragraph{Forward diffusion}
Following \citet{ho2020denoising}, we add a time-dependent Gaussian noise to the logits. %almost-one-hot base logits.%, with the subscript denoting timesteps:
\begin{align}
    \Tilde{\boldsymbol{w}}_0^{c:c+B} &= \operatorname{logits-generation}(\boldsymbol{w}^{c:c+B}) \label{eq:6} \\
    \Tilde{\boldsymbol{w}}_t^{c:c+B} &= \sqrt{\Bar{\alpha}_t} \Tilde{\boldsymbol{w}}_0^{c:c+B} + \sqrt{1-\Bar{\alpha}_t} \boldsymbol{\epsilon}_t \label{eq:7}
\end{align}
where $t \in (1, T)$, $\boldsymbol{\epsilon}_t \sim \mathcal{N}(\boldsymbol{0}, K^2\mathbf{I})$, and $\Bar{\alpha}_t \to 0$ as $t \to T$. At the final step $T$, $\operatorname{softmax}(\Tilde{\boldsymbol{w}}_T^{c:c+B})$ are fully noisy simplexes over $V$, with a 
%standard \han{remove `standard'?} 
logit-normal distribution~\citep{logit-normal}.

\paragraph{Loss function}
In \autoref{eq:2}, a diffusion model is trained to predict the added noise from the noisy representations. Since the forward diffusion process can be computed in a single step (\autoref{eq:1}), the notion here is equivalent to predicting the original data representation \citep{Song2021DenoisingDI,Li2022DiffusionLMIC}. Our objective follows the same intuition but estimates a likelihood instead of the L2 distance while conditioning on additional context:\footnote{L2 distance did not work in our pilot study potentially due to the intrinsically skewed simplex representation.}
\begin{align}
    \mathcal{L}(\theta) &=  \mathbb{E}[
    -\log p_\theta(\boldsymbol{w}^{c:c+B} \mid \Tilde{\boldsymbol{w}}_t^{c:c+B}, \boldsymbol{w}^{<c})] \label{eq:8} \\
    &= \mathbb{E}\left[
    \sum_{j=c}^{c+B-1}-\log p_\theta(w^{j} \mid \Tilde{\boldsymbol{w}}_t^{c:c+B}, \boldsymbol{w}^{<c})\right] \label{eq:9}
\end{align}
% \nonumber\\
$\mathbb{E} [\cdot]$ is a shorthand for $\mathbb{E}_{c \sim \mathcal{U}(1, L-B), t \sim \mathcal{U}(1, T)} [\cdot]$.
The architecture for $\theta$ throughout this work is a bi-directional Transformer encoder~\citep{vaswani2017attention}. Specifically, the input to the model is a concatenation of the context $\boldsymbol{w}^{<c}$ and a sequence of noisy vocabulary simplexes $\operatorname{softmax}(\Tilde{\boldsymbol{w}}_t^{c:c+B})$ of length $B$. The target output is the original tokens $\boldsymbol{w}^{c:c+B}$ at positions $c$ to $c+B$. 

One minimal modification made to the Transformer model is that in addition to the conventional embedding lookup for $\boldsymbol{w}^{<c}$, we modify the embedding layer to take as input a distribution over the vocabulary, $\operatorname{softmax}(\Tilde{\boldsymbol{w}}_t^{c:c+B})$, and compute the embedding vector as a weighted sum of the embedding table. A timestep embedding is also added before the first Transformer block to inform the model of the current timestep.\footnote{More specifically, we have word embeddings for the context, $\operatorname{Emb}_{\text{ctx}}(\boldsymbol{w}^{<c})$, and for the noisy diffusion representations, $W_{\text{diff}} [\operatorname{softmax}(\Tilde{\boldsymbol{w}}_t^{c:c+B})]$. 
The timestep embedding is added to the diffusion word embeddings, $W_{\text{time}}(t/T)$. 
It is similar to positional embeddings, just not varying across sequence positions. We fold it in $\theta$ for notation simplicity.}

\medskip

In \Sref{sec:interpret_training_loss}, we present another interpretation of the training objective as an intuitive contrastive loss.

\subsection{Decoding}
\label{sec:method_decoding}
\paragraph{Logits projection}
Similar to continuous-valued diffusion models, sampling from \methodname{} involves reverse diffusion from $t=T, \ldots, 1$ starting with a Gaussian noise.
At any timestep $t$, our model $\theta$ takes as input noised logits $\tilde{\boldsymbol{w}}_t^{c:c+B}$ and estimates the probability distribution of the original tokens in data by first predicting the logits:
% \begin{align}
%     \boldsymbol{w}_{\text{logits}}^{c:c+B}(\cdot,t) = \operatorname{logits}_\theta(\boldsymbol{w}^{c:c+B} \mid \Tilde{\boldsymbol{w}}_t^{c:c+B}, \boldsymbol{w}^{<c}) \label{eq:13}
% \end{align}
\begin{align}
    \boldsymbol{w}_{\text{logits}, t}^{c:c+B} = \operatorname{logits}_\theta(\boldsymbol{w}^{c:c+B} \mid \Tilde{\boldsymbol{w}}_t^{c:c+B}, \boldsymbol{w}^{<c}) \label{eq:13}
\end{align}
% $p_\theta(\boldsymbol{w}^{c:c+B} \mid \Tilde{\boldsymbol{w}}_t^{c:c+B}, \boldsymbol{w}^{<c})$.
which are then converted to a distribution via softmax. To feed this output to the next step of reverse diffusion, $t-1$, we define a $\operatorname{logits-projection}$ operation to build a predicted data representation close to the initial data representation (almost-one-hot mapping; \autoref{eq:5}). 
We consider three projection operations.
% We experiment with the following projection operations: 
\begin{compactitem}%[leftmargin=*,noitemsep,topsep=0pt,parsep=0pt,partopsep=0pt]
% \vspace{-0.5em}
    \item \emph{Greedy:} creates an almost-one-hot logit centered at the highest probability token.\footnote{This shares a similar intuition as a greedy clamping trick in the embedding-based diffusion in \citet{Li2022DiffusionLMIC}.}  
    % \vspace{-0.5em}
\begin{align}
\hat{w}_{(i)} \text{=} \begin{cases}
+K \text{ if $i$=} \argmax(\boldsymbol{w}_{\text{logits}})\\
-K \text{ otherwise}
\end{cases} \label{eq:14}
\end{align}
% \vspace{-2em}
    \item \emph{Sampling:} creates an almost-one-hot logit centered around a token sampled from the output distribution using top-$p$ sampling~\citep{holtzman2019curious}. $p$ is a hyperparameter.
    % \vspace{-0.5em}
\begin{align}
\hat{w}_{(i)} \text{=} \begin{cases}
+K \text{ if $i$=} \text{top-}p\text{-sample}(\boldsymbol{w}_{\text{logits}})\\
-K \text{ otherwise}
\end{cases} \label{eq:15}
\end{align}
% \vspace{-2em}
    \item \emph{Multi-hot:} creates an almost-one-hot logit centered around \emph{all} tokens in the top-$p$ nucleus.
    % \vspace{-0.5em}
\begin{align}
\hat{w}_{(i)} \text{=} \begin{cases}
+K \text{ if $i \in$ } \text{top-}p\text{-all}(\boldsymbol{w}_{\text{logits}})\\
-K \text{ otherwise}
\end{cases} \label{eq:16}
\end{align}
\end{compactitem}
%The calculation of the top-$p$ vocabulary follows~\citep{holtzman2019curious}. 
%There is no learnable parameter in this projection operation.

\paragraph{Decoding iteration}
Starting from pure noise $\Tilde{\boldsymbol{w}}_T^{c:c+B} \sim \mathcal{N}(\boldsymbol{0}, K^2\mathbf{I})$, in each decoding timestep we compute:
\begin{align}
    &\hat{\boldsymbol{w}}^{c:c+B}_t = \operatorname{logits-projection}(\boldsymbol{w}_{\text{logits}, t}^{c:c+B}) \label{eq:17} \\
    &\Tilde{\boldsymbol{w}}_{t-1}^{c:c+B} = \sqrt{\Bar{\alpha}_{t-1}} \hat{\boldsymbol{w}}^{c:c+B}_t + \sqrt{1-\Bar{\alpha}_{t-1}} \boldsymbol{z} \label{eq:18}
\end{align}
% (\cdot, t)
for $t = T, \ldots, 1$ and $\boldsymbol{z} \sim \mathcal{N}(\boldsymbol{0}, K^2\mathbf{I})$.
% and $\Bar{\alpha}_t \to 1$ as $t \to 0$

At $t=1$, the final $B$-token block is computed simply as $\argmax \Tilde{\boldsymbol{w}}_{0}^{c:c+B}$. To generate the next block, we concatenate the generated block to the previous context to create a new context of length $c+B$ and follow the reverse-diffusion process again as described above. This process can be repeated until the maximum desired length is reached.\footnote{Alternatively, one can also terminate the process if certain special end-of-sequence tokens have been generated.} %repeat this process iteratively to generate longer sequences.  We can concatenate the generation with the context, and repeat the diffusion decoding process for $m$ times. This semi-autoregressive process would generate $m \cdot B$ tokens $\boldsymbol{w}^{c:c+mB}$.

\medskip

It is worth noting that our proposed decoding algorithm is novel and different from the DDPM decoding (\autoref{eq:3}). The DDPM decoding is designed for diffusion in a continuous space and failed to generate sensible outputs in our preliminary experiments based on simplexes. In \Sref{sec:interpret_decoding}, we draw a theoretical connection between our decoding algorithm and DDPM decoding, and also highlight the intuitive difference between the two.

\paragraph{Highly-modular control}
A useful property of continuous diffusion models that naturally arises from their definition is the ability to guide the generated samples to have user-defined attributes at test time. This can be done using gradients from auxiliary models such as classifiers~\citep{Dhariwal2021DiffusionMB}, e.g., guiding the output of an LM to be of a positive sentiment using a sentiment classifier. 
There is a vibrant community of developers on platforms such as HuggingFace where many such text classifiers are publicly available. 
The underlying data representation of \methodname{} is based on vocabulary simplexes. Hence, as long as a classifier shares the same tokenizer as the LM, it can be used for control in an off-the-shelf manner without modifications. This is in contrast to prior work in diffusion language models that do not support such classifiers due to differences in their input representation space \citep{hoogeboom2021argmax,Austin2021StructuredDD,Li2022DiffusionLMIC,Chen2022AnalogBG} and require retraining the classifiers from scratch. 
This ability makes \methodname{} highly modular for controlled text generation and offers key benefits: (1) Training accurate classifiers for many tasks requires huge amounts of data where retraining them can be quite expensive, and (2) this approach allows control from classifiers that are open to use but have been trained on closed source data. 

To guide \methodname{} to generate texts with a target attribute $y$ via a standalone attribute model $f_{\phi}(\cdot)$, we update $\boldsymbol{w}_{\text{logits}, t}^{c:c+B}$ (\autoref{eq:13}) at each timestep $t$ to the form below, drifting according to the gradients from the attribute classifier. 
% Other parts of the decoding algorithm remain unchanged.
\begin{align}
    % \boldsymbol{w}_{\text{logits}}^{c:c+B}(\cdot, t) &\leftarrow 
    % &\boldsymbol{w}_{\text{logits}}^{c:c+B}(\cdot, t) = \\
    % &\boldsymbol{w}_{\text{logits}}^{c:c+B}(\cdot, t) + \lambda \nabla_{\boldsymbol{w}_{\text{logits}}^{c:c+B}} f_{\phi}(y | \boldsymbol{w}_{\text{logits}}^{c:c+B}(\cdot, t), \boldsymbol{w}^{<c}) \label{eq:23}
    &\boldsymbol{w}_{\text{logits}, t}^{c:c+B} + \lambda \nabla_{\boldsymbol{w}_{\text{logits},t}^{c:c+B}} f_{\phi}(y \mid \boldsymbol{w}_{\text{logits}, t}^{c:c+B}, \boldsymbol{w}^{<c}) \label{eq:23}
\end{align}
where $\lambda$ is a hyperparameter balancing the weight of control. The parameters of the standalone attribute model $\phi$ are frozen. We make a trivial modification to the embedding computation as in \Sref{sec:method_training}, to allow the classifier to take as input a simplex.% over the vocabulary.

\subsection{Additional details}
\paragraph{Forward diffusion coefficient $\Bar{\alpha}_t$}
We follow \citet{nichol2021improved} for a cosine schedule of $\Bar{\alpha}_t$:
\begin{align}
    \Bar{\alpha}_t = \frac{r(t)}{r(0)},~r(t) = \cos (\frac{t/T + s}{1 + s}\cdot \frac{\pi}{2})^2
\end{align}
where $s$ is small offset set to 1e-4 in our work and $\alpha_t = \frac{\Bar{\alpha}_t}{\Bar{\alpha}_{t-1}}$.

\paragraph{Fewer timesteps $T$ in decoding}
Decoding from diffusion models requires a series of timesteps ($T$) which can be computationally expensive if $T$ is large. Following \citet{Li2022DiffusionLMIC}, we consider using a smaller value of $T$ at test time to improve decoding speed. 
% the number of decoding steps comparate educing the number of time steps at decoding time does not need as many timesteps as in training.
In this work, we primarily experiment with $T_{\text{decode}} = \frac{T_{\text{train}}}{2}$ and $T_{\text{decode}} = \frac{T_{\text{train}}}{5}$. %, though an even smaller $T_{\text{decode}}$ may also work.

\paragraph{Flexible decoding block size $B$}
Our \methodname{} is trained with a fixed token block size $B_{\text{train}}$. However, the decoding algorithm has a freedom to use a different $B_{\text{decode}}$. In our experiments, we consider both scenarios of $B_{\text{train}} = B_{\text{decode}}$ and $B_{\text{train}} \neq B_{\text{decode}}$. Nevertheless, we leave for future work a more detailed analysis of the impact of the difference between $B_{\text{train}}$ and $B_{\text{decode}}$ on model performance.

\section{Experiments}
\label{sec:experiments}
\subsection{\methodname{} pretraining setup}
\label{sec:experiments_setup}
\paragraph{Model architecture}
% \methodname{} uses a bidirectional Transformer encoder \citep{vaswani2017attention} as the underlying architecture. 
We use a bidirectional Transformer encoder RoBERTa-large \citep{Liu2019RoBERTaAR} (0.4B, comparable size to GPT2-medium) as \methodname{}'s underlying architecture.\footnote{We initialize the model with RoBERTa's weights as well. We observe in our initial exploration that it helps the training loss converge faster than a randomly initialized model. However, given enough computational resources, we conjecture that a randomly initialized model will offer similar performance.} Note that RoBERTa uses a general BPE tokenization \citep{sennrich2016neural}, same as a variety of LMs such as GPT-2 \citep{Radford2019LanguageMA}, GPT-3 \citep{Brown2020LanguageMA}, OPT \citep{Zhang2022OPTOP}, etc. Any attribute classifier using the same tokenization strategy can be used to control \methodname{} in a highly modular way.

\paragraph{Pretraining data, constants, and resource}
We train \methodname{} on the same data as GPT2 to make fair comparisons possible: OpenWebText \citep{Gokaslan2019OpenWeb} which contains 9B tokens. 
% We set the maximum sequence length of the model as 200 tokens. 
Following \citet{Zhang2022OPTOP}, we consider this data as one contiguous sequence of tokens and break it into sequences of length 200 (same as the maximum sequence length our model accepts). We randomly sample 99\% of these sequences for pretraining while leaving the rest as held out for evaluation. We use the following model hyperparameters:\footnote{Future work can do a search given more resources.}
$$L=200, B_{\text{train}}=25, T_{\text{train}}=5000, K=5$$

We use an aggregated batch size of 6,144 and a learning rate of 1e-4 with an AdamW optimizer \citep{Loshchilov2019DecoupledWD}. We trained \methodname{} for 100K steps, which took about 6 days on 32 Nvidia V100 GPUs. %pretraining \methodname{} for 100k steps.

\paragraph{Pretraining loss}
Canonical training-time perplexity of LMs is not compatible with diffusion LMs due to the difference in the inputs to the models (\autoref{eq:4} and \autoref{eq:9}). 
Our pretraining loss is a per-token negative log-likelihood (NLL) that depends on the specific noise schedule being used. \methodname{} gets an average NLL of 3.87 at the end of pretraining. We show a pretraining loss curve in the appendix (\Sref{sec:appendix_results}).

\subsection{Unconstrained text generation}
\label{sec:experiments_natural_gen}

\begin{table*}[t]
    \centering
    \begin{tabular}{@{}lp{0.60in}p{0.35in}p{0.42in}p{0.40in}p{0.40in}p{0.40in}p{0.32in}p{0.32in}@{}}
    \toprule
     (Length 50) & MAUVE $\uparrow$ & PPL \small{$\xrightarrow[\text{gold}]{}$} & \small{$|\Delta_{\log \text{PPL}}|$} \normalsize{$\downarrow$} & Dist-1 $\uparrow$ & Dist-2 $\uparrow$ & Dist-3 $\uparrow$ & Zipf \small{$\xrightarrow[\text{gold}]{}$} & Rep $\downarrow$ \\
      \midrule
      \textit{Gold continuation} & 100.00 & 17.75  & 0.00 & 88.62 & 95.88 & 93.71  & 0.88 & 0.10   \\
      [2pt]
      \underline{GPT2-medium} \tiny{(Best config)} \\
      \small{Top-$p$=0.95} & 96.57 \qquad \tiny{$\pm$ 0.40} & 12.72 \qquad \tiny{$\pm$ 0.07} & 0.33 & 66.31 \qquad \tiny{$\pm$ 0.11} & 91.77 \qquad \tiny{$\pm$ 0.03} & 92.75 \qquad \tiny{$\pm$ 0.06} & 1.01 & 0.26 \qquad \tiny{$\pm$ 0.04}  \\
      \underline{GPT2-large} \tiny{(Best config)} \\
      \small{Top-$p$=0.95} & 96.41 \qquad \tiny{$\pm$ 0.78} & 10.57 \qquad \tiny{$\pm$ 0.05} & 0.51 & 64.91 \qquad \tiny{$\pm$ 0.13} & 90.88 \qquad \tiny{$\pm$ 0.06} & 92.38 \qquad \tiny{$\pm$ 0.05} & 1.01 & 0.41 \qquad \tiny{$\pm$ 0.06}  \\
      \underline{GPT2-xl} \tiny{(Best config)} \\
      \small{Typical-$\tau$=0.95} & 97.03 \qquad \tiny{$\pm$ 0.50} & 10.33 \qquad \tiny{$\pm$ 0.04} & 0.54 & 64.87 \qquad \tiny{$\pm$ 0.15} & 90.69 \qquad \tiny{$\pm$ 0.07} & 92.16 \qquad \tiny{$\pm$ 0.05} & 1.01 & 0.37 \qquad \tiny{$\pm$ 0.04}  \\
      [0pt]
      \underline{\methodname{}-``medium''} \tiny{(Top-3)} \\
      [0pt]
      \small{Sampling $p$=0.99, $T$=1000} & \textbf{97.89} & 30.68 & 0.54 & \textbf{68.99} & \textbf{92.60} & \textbf{92.94} & 1.01 & \textbf{0.16}  \\
      [0pt]
      \small{Sampling $p$=0.95, $T$=1000} & 96.64 & 27.34 & 0.43 & 67.75 & 92.16 & 92.91 & 1.01 & \textbf{0.16}  \\ 
      [0pt]
      \small{Sampling $p$=0.9, $T$=2500} & 96.46 & 20.56 & \textbf{0.14} & 66.61 & 91.46 & 92.56 & 1.05 & 0.26  \\
      \bottomrule
    \end{tabular}
    \caption[Caption for LOF]{Unconstrained generation evaluation of \methodname{} and GPT-2 models at length 50. %Indication of better quality: MAUVE $\uparrow$, PPL $\rightarrow$ PPL$_{\text{gold}}$, Dist-1/2/3 $\uparrow$, Zipf $\rightarrow$ Zipf$_{\text{gold}}$, Rep $\downarrow$, BLEU $\uparrow$. 
    % PPL is computed with GPT-Neo-1.3B \citep{gpt-neo}. 
    For GPT-2 models, the results are averaged across 5 random seeds, and we show the best sampling parameter configuration. For our \methodname{}, we show the top-3 configurations. All configurations are ranked based on MAUVE, with original parameters from \citet{Pillutla2021MAUVEMT}. The perplexity (PPL) is measured by GPT-Neo-1.3B.\protect\footnotemark
    % \han{EXTRA RESULTS (non-total PPL), from gold to gpt2 medium/large/xl to our three configs: 21.77, 14.27, 11.70, 11.54, 41.17, 36.10, 25.95}
    }
    \label{tab:main_result}
    \vspace{-0.8em}
\end{table*}

\paragraph{Setup}
% Our first main evaluation in this work is
First, we benchmark \methodname{} with autoregressive LMs trained on the same data (GPT2) on text generation quality. We randomly sample 1000 sequences from the held-out OpenWebText test data, extract their prefixes as prompts (context), and generate continuations from the LMs.
%use prefixes of the text strings as contexts to the LMs, and let the models generate a continuation. 
We consider three setups: with prompt lengths 25, 50 and 100 with respective output lengths as 25, 50 and 100 tokens.  In each setup, we sample 5 continuations for each input context, thus comparing the quality of 5,000 generations from baseline GPT-2 models and our \methodname{}.

We compare \methodname{} with  GPT2-medium, large and xl models (containing 0.4B, 0.8B and 1.6B parameters respectively) as baselines. For reference, our model size is comparable to GPT2-medium. We experiment with two popular decoding strategies for the baseline GPT-2 models with canonical parameters: nucleus sampling \citep{holtzman2019curious} with a top-$p$ of 0.9 and 0.95, and typical sampling \citep{Meister2022LocallyTS} with a typical-$\tau$ of 0.2 and 0.95.

\footnotetext{
MAUVE, Dist-1/2/3, and Rep are in percentage. PPL is obtained through a micro average following \citet{holtzman2019curious,Pillutla2021MAUVEMT,Meister2022LocallyTS}. 
% Decoding from \methodname{} is expensive and thus not experimented with 5 random seeds given the number of configurations.
}

For \methodname{}, we consider three logits projection strategies, sampling and multi-hot with $\text{top-}p \in \{ 0.0, 0.1, 0.2, 0.5, 0.7, 0.9, 0.95, 0.99 \}$, and greedy (which is functionally equivalent to the sampling with top-$p$=0). We use a test block size ($B_\text{decode}$) of 25. When generating samples of length 50 or 100, we semi-autoregressively sample in blocks of 25 and feed them as additional context to generate the next block as described in \Sref{sec:method_decoding}. 

We evaluate the generated continuations on two axes: quality and diversity. As automatic quality metrics, we report perplexity measured by a separate, larger language model ~\citep[GPT-Neo-1.3B,][]{gpt-neo}. Prior works, however, have shown that low perplexity of generated text is not necessarily an indication of high quality but of degenerate behavior~\citep{nadeem2020systematic,zhang2021trading} and have proposed closeness to the perplexity of human-written text as a better evaluation. Hence, we also report the difference of log perplexity between the generated text and human-written continuations ($|\Delta_{\log \text{PPL}}|$). For diversity evaluation, we report Zipf's coefficient (Zipf) and average distinct $n$-grams in the output samples \citep[Dist-$n$]{li-etal-2016-diversity}. In addition, we also report the repetition rate \citep[Rep]{welleck2020neural,holtzman2019curious}, measuring the proportion of output samples that end in repeating phrases. Finally, we report MAUVE~\citep{Pillutla2021MAUVEMT} which evaluates both quality and diversity together by approximating information divergence between generated samples and human-written continuations (from the OpenWebText held-out set).

% \begin{figure*}[t]
%     \centering
%     \begin{subfigure}[h]{0.32\textwidth}
%         \includegraphics[width=\textwidth]{resources/mauve.png}
%         % \caption{TBD}
%         \label{fig:1-1}
%     \end{subfigure}
%     \hfill
%     \begin{subfigure}[h]{0.32\textwidth}
%         \includegraphics[width=\textwidth]{resources/ppl.png}
%         % \caption{TBD}
%         \label{fig:1-2}
%     \end{subfigure}
%     \hfill
%     \begin{subfigure}[h]{0.32\textwidth}
%         \includegraphics[width=\textwidth]{resources/rep.png}
%         % \caption{TBD}
%         \label{fig:2-2}
%     \end{subfigure}

%     % \vfill

%     % \begin{subfigure}[h]{0.32\textwidth}
%     %     \includegraphics[width=\textwidth]{resources/zipf.png}
%     %     % \caption{TBD}
%     %     \label{fig:2-1}
%     % \end{subfigure}
%     % \hfill
%     % \begin{subfigure}[h]{0.32\textwidth}
%     %     \includegraphics[width=\textwidth]{resources/dist.png}
%     %     % \caption{TBD}
%     %     \label{fig:1-3}
%     % \end{subfigure}
%     % \hfill
%     % \begin{subfigure}[h]{0.32\textwidth}
%     %     \includegraphics[width=\textwidth]{resources/bleu.png}
%     %     % \caption{TBD}
%     %     \label{fig:2-3}
%     % \end{subfigure}
%     \caption{Influence of different decoding logits projection strategies and associating top-$p$ for \methodname{} on various text quality metrics. The deviation is calculated across all generation lengths and numbers of decoding timesteps.}
%     \label{fig:6_vis}
% \end{figure*}

\paragraph{Results}
\autoref{tab:main_result} summarizes our main results on the 50-token prompt and output setup. We report the numbers for the best performing three settings for logits projection and decoding steps $T$ in \methodname{}. We report the best setting for the baselines. The results for other generation lengths have a similar trend and can be found in the appendix (\Sref{sec:appendix_results}).

% \textbf{\methodname{} generates better quality and more diverse texts than GPT-2.} 
We find that \methodname{}, though being smaller in size, outperforms larger GPT-2 models on the unified metric MAUVE. On diversity, \methodname{} outperforms GPT-2 in Dist-$n$ while achieving lower repetition rates. % The repetition rate of \methodname{} is also lower than GPT-2 at all sizes. 
% We additionally find the BLEU score to gold continuation higher in \methodname{} compared to the same-size GPT-2 model. 
On perplexity, the results are slightly mixed. We observe a trade-off between MAUVE and perplexity for different settings we considered, 
% underperforms GPT2-xl on perplexity (in terms of closeness to gold PPL; see $|\Delta_{\log \text{PPL}}|$)
% Our approach, in some settings, outperforms the baselines (in terms of closeness to gold PPL; see $|\Delta_{\log \text{PPL}}|$), while lags behind in the others, 
indicating that further tuning of the hyperparameters may be required. 
However, one of our best performing settings (sampling top-$p$=0.9, $T$=2500) still achieves the closest perplexity to the gold continuation. 
% Overall, our best performing setting (Sampling $p=0.9, T=2500$) matches or outperforms both GPT2-medium and large on \textit{all} metrics. We hypothesize the gap from GPT2-xl will be filled with larger versions of \methodname{}. We leave this direction for future work. 

%  Nevertheless, 

%The GPT-2 generations' perplexity is lower than \methodname{}, but lower perplexity does not necessarily indicate better quality \citep{nadeem2020systematic,zhang2021trading}. 
%We measure the difference of log-perplexity between the model generations and the gold continuations, and find \methodname{} has a smaller difference with certain configurations.

% \autoref{fig:6_vis} shows the influence of different logits projection strategies and the associated parameters on the output text quality. We observe that reducing top-$p$ $\to$ 0 (greedy projection) can lead to a low perplexity but it is undesirable due to a high repetition rate. We also find the multi-hot projection strategy is overall worse performing than the sampling projection strategy in our setup, indicating it is better to commit the intermediate states to single rather than multiple tokens. 
% This can be because our logits mapping involves putting probability mass on singular tokens. The multi-hot projection may still be a viable strategy if future work uses multi-hot logits mapping for the input tokens. 

% moving above to appendix
In \Sref{sec:appendix_results}, we show the influence of different logits projection strategies and the associated parameters on the output text quality in \autoref{fig:6_vis}. We also show qualitative examples of the generations by \methodname{} in \autoref{tab:qual_ex} and a trajectory of intermediate states during the decoding process in \autoref{tab:traj_qual_ex}. 

% This is an expected result since our training process involves putting probability mass on singular tokens. So at decoding time it is better to commit a single rather than multiple tokens in intermediate states. However, in future work adding noise which allow multi-token peaks may be considered making this it is viable decoding strategy. 
% the intermediate states within the diffusion decoding process to a single token rather than should commit to a single rather than a set of tokens in the vocabulary.

% Finally, we show a few qualitative examples of the generations by \methodname{} in \autoref{tab:qual_ex} and a trajectory of intermediate states during the decoding process in \autoref{tab:traj_qual_ex} (\Sref{sec:appendix_results}).

%%%%%%%%%%%%%%%% NEW COMPARISON %%%%%%%%%%%%%%%%

\begin{table}[t]
    \centering
    \begin{tabular}{@{}lp{0.6in}p{0.6in}@{}}
    \toprule
      (ROCStories) & MAUVE & PPL  \\
      \midrule
      \textit{Gold continuation} & 100.00 & 18.57   \\
      [1pt]
      \underline{Diffusion-LM} & 46.11 & 35.96   \\
    %   Diffusion-LM \tiny{\han{BERT-init, maybe remove}} & 0.4 & 73157.7   \\
      [1pt]
      \underline{\methodname{}} & \textbf{87.22}  & \textbf{22.91}  \\ 
      \bottomrule
    \end{tabular}
    \caption[Caption for LOF]{
    Unconstrained generation results of \methodname{} and Diffusion-LM on ROCStories with 50 prompt tokens and 50 output tokens. We report the MAUVE score between the gold continuation and model generations. 
    We also show the perplexity (PPL) of model generations measured by GPT-Neo-1.3B.\protect\footnotemark
    }
    \label{tab:comparison_with_difflm}
    \vspace{-0.8em}
\end{table}
\footnotetext{Due to a lowercase tokenization of ROCStories, we use BERT-base-uncased as MAUVE's embedding model here.}

\paragraph{Comparison with \citet{Li2022DiffusionLMIC}} 
% So far we have pretrained a simplex-based diffusion language model on established pretraining corpora (e.g., OpenWebText) and benchmarked it with standard autoregressive language models (e.g., GPT-2). 
A prior work to us, \citet{Li2022DiffusionLMIC} propose Diffusion-LM, an embedding-based diffusion model trained on two small toy datasets, E2E \citep{novikova2017e2e} and ROCStories \citep{mostafazadeh2016corpus}. 
In this subsection, we make a diversion to compare the embedding-based Diffusion-LM with our semi-autoregressive, simplex-based \methodname{}. 
Following \citet{Li2022DiffusionLMIC}, we train a Diffusion-LM on ROCStories with a default embedding size of 128, 0.1B parameters under a BERT-base \citep{Devlin2019BERTPO} structure,\footnote{We train two versions of Diffusion-LM, with and without BERT's encoder weights as an initialization. The default no-initialization setup as in \citet{Li2022DiffusionLMIC} works reasonably, while the other degenerates. Details can be found in \Sref{sec:appendix_comparison_difflm}.} and a sequence length of 100.  
For a fair comparison, \emph{only within this subsection} we train a \methodname{} with ROCStories sequences of 100 tokens, a decoding block size of 25, and a BERT-base initialization. Further details of the setup can be found in \Sref{sec:appendix_comparison_difflm}. 

On 2,700 held-out ROCStories sequences, we use the first 50 tokens of each sequence as a prompt and have the model generate the next 50. 
In \autoref{tab:comparison_with_difflm}, we show the MAUVE score and perplexity of both models. We observe a substantially higher MAUVE score and lower perplexity with \methodname{}.

%%%%%%%%%%%%%%%%%%%%%%%%%%%%%%%%%

\subsection{Controlled text generation}
\label{sec:experiments_controlled_gen}
\paragraph{Setup}
% \methodname{} enables fully modular control by external attribute classifiers that share a same tokenization method. 
To evaluate \methodname{}'s ability for highly-modular control, we consider the task of sentiment controlled generation where given a prompt, the goal is to generate a continuation with a positive (or negative) polarity. 
% We evaluate the model's ability to generate texts with specific sentiments, a canonical shared task in multiple prior work. 
We use a set of 15 short prompts as in  \citet{dathathri2019plug} and generate 20 samples per prompt per sentiment category, making the total number of generated samples to be 600. Following~\citet{mireshghallah2022mix}, we generate samples with 3 different output lengths: 12, 20 and 50. 
% making 600 generations (2 sentiments, 15 prompts, 20 samples for each sentiment and prompt). Experiments are performed at a length of 12, 20, and 50 tokens.
% To generate texts with specific sentiment using \methodname{},
For guidance, we simply import a popular sentiment classifier\footnote{\url{https://huggingface.co/cardiffnlp/twitter-roberta-base-sentiment}} from HuggingFace trained with Twitter sentiment data with over 58M training examples~\citep{barbieri2020tweeteval}. This model serves as $f_{\phi}(\cdot)$ as shown in \autoref{eq:23}. 
%This is our \emph{internal} classifier for guiding the decoding, for which it is easy to obtain a high accuracy. 
In addition to quality and diversity of the generated samples, we also evaluate them on control (that is measuring if the generated output is actually positive or negative in polarity). For this, we use an \emph{external} sentiment classifier trained on a different dataset. Specifically, we use a classifier trained with Yelp reviews\footnote{\url{https://huggingface.co/textattack/bert-base-uncased-yelp-polarity}} \citep{zhang2015character,morris2020textattack} following the evaluation setup in the baselines we consider.

% For the \methodname{} decoding hyperparameters, 
Again, we consider the sampling and multi-hot decoding strategies with $\text{top-}p \in \{ 0.2, 0.5, 0.9 \}$, $T_{\text{decode}} \in \{ 1000, 2500, 5000 \}$, and the multiplier for control $\lambda \in \{ 0, 100, 500, 2000 \}$. For the generation of 12/20/50 tokens, we use $B_{\text{decode}}$=12/20/25 and apply the decoding algorithm for $m$=1/1/2 iterations respectively.

\begin{table}[t]
    \centering
    \begin{tabular}{@{}p{0.9in}p{0.58in}p{0.38in}p{0.68in}@{}}
    \toprule
      (Length 50) & C-Ext.\tiny{(Int.)} & PPL & Dist-1/2/3 \\
      \midrule
      \underline{DAPT}$^{\mathbb{CM}}$ & 79.8 & 57.2  & 61/92/94 \\
      [2pt]
      \underline{PPLM}$^{\mathbb{CC}}$ & 60.7 \tiny{(73.6)} & 29.0  & - \\
      [2pt]
      \underline{FUDGE}$^{\mathbb{CC}}$ & 59.1 & \textbf{8.4}  & 47/83/92 \\
      [2pt]
      \underline{GeDi}$^{\mathbb{CM}}$ & \textbf{99.2} & 107.3  & 71/93/92 \\
      [2pt]
      \underline{DExperts}$^{\mathbb{CM}}$ & 94.8 & 37.1  & 56/90/92 \\
      [2pt]
      \underline{MuCoLa}$^{\mathbb{CC}}$ & 86.0 & 27.8  & 52/76/80 \\
      \midrule
      [2pt]
      \underline{M\&M LM$^{\mathbb{HMC}}$} & 68.6 \tiny{(93.8)} & 122.3  & - \\
      [2pt]
      \underline{\methodname{}$^{\mathbb{HMC}}$} & \textit{94.1} \tiny{(99.0)} & \textit{23.1}  & 46/84/92 \\ % total ppl 20.5
      \bottomrule
    \end{tabular}
    \caption[Caption for LOF]{Controlled text generation results of \methodname{} and baselines at length 50. We report the external classifier's accuracy (C-Ext.) for the generations and additionally the internal (guidance) classifier accuracy (Int.) if available. The perplexity (PPL) is computed with GPT2-xl. MuCoLa is the version using two discriminators. 
    $\mathbb{CM}$ stands for customized language model, $\mathbb{CC}$ stands for customized classifier, and $\mathbb{HMC}$ stands for highly-modular classifier (in an order of increasing modularity). The best of all results are boldfaced, and the best of $\mathbb{HMC}$ results are italicized.\protect\footnotemark 
    }
    \label{tab:ctr_result_3}
    \vspace{-0.8em}
\end{table}

\paragraph{Results}
We show the quality of the controlled generations from three perspectives: target attribute via the external classifier accuracy, fluency via perplexity, and diversity via the distinctiveness measures. In \autoref{tab:ctr_result_3}, we show the experimental results for output length 50. The results at length 12 and 20 have a similar trend and can be found in the appendix (\Sref{sec:appendix_results}).

Among the baseline methods, DAPT \citep{Gururangan2020DontSP}, GeDi \citep{krause2021gedi}, and DExperts \citep{liu2021dexperts} require training customized language models aware of the desired attributes (denoted as CM in \autoref{tab:ctr_result}). PPLM \citep{dathathri2019plug}, FUDGE \citep{yang2021fudge}, and MuCoLa \citep{Kumar2022ConstrainedSF} require training a customized attribute classifier (CC). While our proposed method \methodname{} and M\&M LM \citep{mireshghallah2022mix} can directly import mainstream existing attribute classifiers from platforms like HuggingFace and are thus highly modular (HMC). We show the baseline results as reported in \citet{mireshghallah2022mix} and \citet{Kumar2022ConstrainedSF}.

\methodname{} shows strong controllability while possessing great modularity. \methodname{} outperforms M\&M LM, the other HMC method by a large margin. Even when comparing with the CC and CM methods, our method achieves a good balance in control, fluency, and diversity.

\footnotetext{
PPL is obtained through a macro average following \citet{Kumar2022ConstrainedSF}.
}

% \autoref{fig:2_vis} shows the impact of the control weight $\lambda$ and top-$p$ on the attribute accuracy and perplexity. As expected, a larger control weight leads to a better external classifier accuracy. The perplexity at the same time increases with a larger $\lambda$, but under a reasonable range for a top-$p$ of 0.2 and 0.5. 
% We show qualitative examples of the controlled generations by \methodname{} in \autoref{tab:qual_ex} (\Sref{sec:appendix_results}).

In \Sref{sec:appendix_results}, we show the impact of the control weight $\lambda$ and top-$p$ on the attribute accuracy and perplexity in \autoref{fig:2_vis}. We also show qualitative examples of the controlled generations by \methodname{} in \autoref{tab:qual_ex}.

\section{Related work}
\label{sec:related_work}

\paragraph{Diffusion models}
Diffusion models have demonstrated impressive performance in popular continuous-valued domains such as images~\citep{ho2020denoising}, audio~\citep{Kong2021DiffWaveAV}, video~\citep{Ho2022VideoDM} and recently also been adopted for 3D-shapes, protein structures, and more~\citep{zhou20213d,Trippe2022DiffusionPM,Wu2022ProteinSG}. Since they are based on adding Gaussian noise, these approaches are not straightforward to apply to discrete valued domains like text. 
\citet{hoogeboom2021argmax,Austin2021StructuredDD} propose diffusing in the discrete space using categorical distributions which are modified using transition matrices. However, these methods do not straightforwardly support control and yield worse results than comparable autoregressive models. 
\citet{Li2022DiffusionLMIC} propose to represent each token as a continuous embedding and apply diffusion in the embedding space. 
%where they apply several heuristics to learn the embedding table together with the model. 
They train the LM to generate a fixed length sequence whereas \methodname{} allows flexibility in the generated sequence length by generating block-wise. Further, their LM is trained with specialized datasets and not evaluated against general-purpose autoregressive LMs on unconstrained text generation. Their method supports post-hoc control but requires training a customized attribute classifier,\footnote{The control for diffusion models can also be classifier-free \citep{ho2021classifier} but requires training with the target attribute in advance, which is not a focus of this work.} since the diffusion operates on a learned embedding space. 
\citet{Gong2022DiffuSeqST}, a concurrent work to ours, extend \citet{Li2022DiffusionLMIC} to a sequence-to-sequence setup with a similar underlying embedding-based method. 
Our work is most closely related to \citet{Chen2022AnalogBG} which transform discrete data into a sequence of bits and represent each bit as +1 or -1 converting it into a continuous-valued domain. For textual data, however, it can lead to extremely long sequences which are difficult to optimize. In this work, we instead maintain a subword based vocabulary but represent each token as a sequence of manually defined logits. 

\paragraph{Language models}
The majority of existing language models for text generation are trained autoregressively, i.e., they predict the next token given previously generated context. This paradigm scaled up both in terms of model size and training data size has resulted in impressive capabilities on many benchmarks~\citep{Brown2020LanguageMA,Chowdhery2022PaLMSL}. However, they generate text one token at a time which does not provide flexible control over attributes of the generated text. Non-autoregressive models which generate the entire output sequence at the same time have also been explored in prior work other than diffusion models~\citep{Lee2018DeterministicNN,Ghazvininejad2019MaskPredictPD}. However, they are primarily focused on improving decoding efficiency and applied for specialized tasks like translation~\citep{gu2018non,kaiser2018fast,wang2019non} and text editing~\citep{Gu2019LevenshteinT}. Many of these work have iterative processes in a discrete space, with some exploring continuous representations \citep{ma2019flowseq,lee2020iterative}. 
To address the quality decline with the non-autoregressive methods compared to autoregressive models, prior work have also explored semi-autoregressive approaches~\citep{wang2018semi,qi2021bang}. In the same vein, our work seeks to address the drawbacks of autoregressive language models and non-autoregressive diffusion models with a middle ground. 

\paragraph{Controllable text generation}
Early solutions for controlling attributes of generated text focused on training or finetuning AR-LMs with specific control codes~\citep{keskar2019ctrl,Gururangan2020DontSP,chan2021cocon}. These methods are difficult to extend to new controls as it requires retraining the models. 
More recent work includes decoding approaches from pretrained AR-LMs without modifying the models, through altering the output probability distribution at each step using different control objectives~\citep{dathathri2019plug,krause2021gedi,yang2021fudge,liu2021dexperts,lu-etal-2021-neurologic,pascual-etal-2021-plug-play}. However, these methods do not allow modifying a token once it is generated and are thus suboptimal for controls at the scope of the whole sequence. Closely related to \methodname{} are \citet{kumar2021controlled,Qin2022COLDDE,Kumar2022ConstrainedSF}, which propose gradient-based decoding algorithms from AR-LMs. They require computing a backward pass through the LMs for each iteration, an expensive operation. In contrast, \methodname{} with its semi-autoregressive setup allows editing past tokens via diffusion. In addition, most of these approaches require training control functions from scratch whereas our model allows using off-the-shelf classifiers. 
\citet{mireshghallah2022mix} propose a non-autoregressive LM based on Metropolis-Hastings sampling. It also supports off-the-shelf classifiers for control, and we therefore use it as a direct baseline for \methodname{}.

\section{Conclusion}
We present \methodname{}, a semi-autoregressive diffusion based language model trained to denoise corrupted simplexes over the output vocabulary. Compared to prior work in text-based diffusion, \methodname{} offers more flexibility in output length by generating blocks of text and an ability to use off-the-shelf attribute classifiers for control without additional tuning. 
%(simplex-based). 
On unconstrained text generation, \methodname{} performs on par with or outperforms strong and larger autoregressive baselines (GPT-2) in generation quality and diversity, while vastly outperforming diffusion baselines (Diffusion-LM). On controlled text generation, \methodname{} surpasses baselines while possessing an easy-to-use modular design. 
We believe that  \methodname{} opens an exciting direction for future research in flexible and modular diffusion-based language generation.

\section*{Limitations}
\paragraph{Sample efficiency}
In AR-LMs, an NLL loss is computed at training time for every token in the sequence of length $L$   (\autoref{eq:4}). However, in \methodname{}, each time a pretraining example is sampled, the loss is computed on only $B$ tokens (\autoref{eq:9}) leading to a lower sample efficiency than AR-LM. Towards improving this efficiency, future work could explore model architectures dedicated to semi-autoregressive diffusion rather than the vanilla Transformer encoder we use in this work.

\paragraph{Decoding speed}
Since each block is generated by refining over several iterations, \methodname{} has a considerably slower decoding speed than autoregressive models.
% at generating samples than autoregressive models.
% \emph{s}uper \emph{s}low \emph{d}ecoding speed compared to autoregressive LMs. 
For example, given a context of 50 tokens (single instance, unbatched), it takes \methodname{} 25 seconds to generate the next block of 25 tokens ($T_{\text{decode}}$=1000). While our work focused on establishing the efficacy of diffusion-based LMs and modular controlled generation, future work could explore tuning $T_{\text{decode}}$ to balance model performance and decoding speed, or more efficient training and decoding algorithms extending ideas from prior work on diffusion models for continuous domains~\citep{Song2021DenoisingDI,nichol2021improved,Rombach2022HighResolutionIS,Meng2022OnDO}.

\paragraph{Decoding block size}
In this work, although we allow setups where $B_{\text{train}} \neq B_{\text{decode}}$, the decoding block size $B_{\text{decode}}$ remains the same across $m$ decoding iterations, leaving space for a more flexible decoding schedule. Future work can also explore learning $B_{\text{decode}}$ (and $B_{\text{train}}$) rather than using constant pre-defined lengths.

\medskip

Larger scale experiments with different kinds of controls and their combinations can be done, as well as more sophisticated ways to incorporate them~\citep{kumar2021controlled}. In addition, we plan to explore alternative methods to continuously represent and add noise to discrete text~\citep{Bakosi2013ASD}. 
This work experiments with pretraining data that is primarily in English. Future work can also explore challenges and benefits of diffusion-based LMs in a multilingual setup.

\section*{Ethics statement}
Language models trained on data from the web can perpetuate social biases and toxic interactions, and can be prone to generating harmful language \citep{gehman-etal-2020-realtoxicityprompts,wallace-etal-2019-universal,wallace-etal-2020-imitation,sheng-etal-2021-societal,10.1145/3531146.3533088}. 
Further, language generation models could memorize and amplify patterns in data without deeper language understanding or control, so they can be factually inconsistent and generate disinformation \citep{factentailment, pagnoni2021understanding, zellers2019defending}, or can compromise user privacy \citep{DBLP:conf/uss/CarliniTWJHLRBS21}. 
Prior works have outlined these risks~\citep{sheng-etal-2021-societal, DBLP:journals/corr/abs-2112-04359}, discussed their points of origin, and advocated for future research on ethical development of LMs~\citep{10.1145/3442188.3445922,solaiman2019release}. 

%\citep{zellers2019neuralfakenews, pagnoni2021understanding}. 
While these studies have been conducted for autoregressive LMs, our diffusion-based LM is subject to these problems as well. However, since our method naturally incorporates controllability, future work may explore control functions that could potentially alleviate these issues~\citep{liu2021dexperts,Kumar2022ConstrainedSF}. 
One risk is that controllability can also be misused maliciously, with 
% The key risk of controllability is dual use: potential for malicious uses, 
models being intentionally exploited to generate biased, toxic, or non-factual content \citep{bagdasaryan2022spinning,pagnoni2022threat}. Therefore, apart from controlled generation, future work should aim to detect the generations under control as well to defend against the malicious use~\citep{kumar2022language}.

\section*{Acknowledgements}
The authors would like to thank Tianxiao Shen, Tianxing He, Jiacheng Liu, Ruiqi Zhong, Sidney Lisanza, Jacob Gershon, members of TsvetShop, and the anonymous ACL reviewers for their helpful discussions and feedback. 
X.H.~gratefully acknowledges funding from the UW-Meta AI Mentorship program. S.K. gratefully acknowledges a Google Ph.D. Fellowship. Y.T.~gratefully acknowledges an Alfred P.~Sloan Foundation 
Fellowship. 
This research is supported in part by by the National Science Foundation (NSF) under Grants No.~IIS2203097, IIS2125201, and NSF CAREER Grant No.~IIS2142739. 
This research is supported in part by the Office of the Director of National Intelligence (ODNI), Intelligence Advanced Research Projects Activity (IARPA), via the HIATUS Program contract \#2022-22072200004. The views and conclusions contained herein are those of the authors and should not be interpreted as necessarily representing the official policies, either expressed or implied, of ODNI, IARPA, or the U.S. Government. The U.S. Government is authorized to reproduce and distribute reprints for governmental purposes notwithstanding any copyright annotation therein.

%Below copied from previous work
%This material is based upon work supported by NSF grants IIS1812327 and SES1926043, and by Amazon MLRA award.
%Wallace's contributions were supported by the Army Research Office (W911NF1810328).
%We also thank Amazon for providing GPU credits.

% \medskip

\bibliography{my_cites}
\bibliographystyle{acl_natbib}

\appendix

% \section{Example Appendix}
% \label{sec:appendix}

% This is an appendix.

\section{A contrastive interpretation of the training loss} 
\label{sec:interpret_training_loss}

The training of \methodname{} is simply maximizing the likelihood $\log p_\theta(\boldsymbol{w}^{c:c+B} \mid \Tilde{\boldsymbol{w}}_t^{c:c+B}, \boldsymbol{w}^{<c})$. 
This diverts from the exact objective of DDPM that is supported by a variational bound. 
% Though it is inheriting techniques of diffusion models which justifies the objective from a perspective of variational bound,
However, below we give an intuitive interpretation to our objective.
\par\nopagebreak\noindent\ignorespaces
{\small
\begin{align}
    &\log p_\theta(\boldsymbol{w}^{c:c+B} \mid \Tilde{\boldsymbol{w}}_t^{c:c+B}, \boldsymbol{w}^{<c})\\
    =&\log \frac{p_\theta(\boldsymbol{w}^{c:c+B} \mid \boldsymbol{w}^{<c})~p_\theta(\Tilde{\boldsymbol{w}}_t^{c:c+B} \mid \boldsymbol{w}^{c:c+B}, \boldsymbol{w}^{<c})}{p_\theta(\Tilde{\boldsymbol{w}}_t^{c:c+B} \mid \boldsymbol{w}^{<c})}\\
    =&~\log \underbrace{\textstyle p_\theta(\boldsymbol{w}^{c:c+B} \mid \boldsymbol{w}^{<c})}_{\mathclap{\text{\scriptsize likelihood of true data}}} - \log \underbrace{\textstyle p_\theta(\Tilde{\boldsymbol{w}}_t^{c:c+B} \mid \boldsymbol{w}^{<c})}_{\mathclap{\text{\scriptsize likelihood of noisy data at timestep }t}} \nonumber \\
    &+ \log \underbrace{\textstyle p(\Tilde{\boldsymbol{w}}_t^{c:c+B} \mid \boldsymbol{w}^{c:c+B})}_{\mathclap{\text{\scriptsize forward diffusion process independent of }\theta}}
\end{align}
}%
Optimizing $\theta$ is a contrastive objective: maximizing the estimated likelihood of true data, while penalizing the estimated likelihood of noisy data under a broad range of different noise scales.

\section{Connection between our decoding algorithm and the DDPM decoding}
\label{sec:interpret_decoding}
We revisit the decoding step in DDPM introduced in \autoref{eq:3}. Since we know that during the training phase $\boldsymbol{x}_{t}$ is generated through a one-step forward diffusion process (\autoref{eq:1}), a model $\theta$ predicting the added noise $\epsilon_{\theta}(\boldsymbol{x}_t, t)$ can therefore be considered as predicting an imaginary $\boldsymbol{x}_0$ in one-step:
\begin{align}
    \hat{\boldsymbol{x}}_0(\boldsymbol{x}_t, t, \theta) &= \frac{1}{\sqrt{\Bar{\alpha}_t}}(\boldsymbol{x}_t - \sqrt{1-\Bar{\alpha}_t} \epsilon_{\theta}(\boldsymbol{x}_t, t)) \label{eq:19}
    % \boldsymbol{x}_t &= \sqrt{\Bar{\alpha}_t}\hat{\boldsymbol{x}}_0(\boldsymbol{x}_t, t, \theta) +  \sqrt{1-\Bar{\alpha}_t} \epsilon_{\theta}(\boldsymbol{x}_t, t)  \label{eq:19x}
\end{align}
Below we write $\hat{\boldsymbol{x}}_0(\boldsymbol{x}_t, t, \theta)$ as $\hat{\boldsymbol{x}}_0$ and $\epsilon_{\theta}(\boldsymbol{x}_t, t)$ as $\epsilon_{\theta}$ for simplicity.

Rearranging the DDPM decoding transition (\autoref{eq:3}), we have:
\begin{align}
    \boldsymbol{x}_{t-1} 
    % &= \frac{1}{\sqrt{\alpha_t}} (\boldsymbol{x}_{t} - \frac{1-\alpha_t}{\sqrt{1-\Bar{\alpha}_t}} \epsilon_{\theta}) \label{eq:20} \\
    % &= (\sqrt{\Bar{\alpha}_{t-1}}\hat{\boldsymbol{x}}_0 + \frac{\alpha_t - \Bar{\alpha}_t}{\sqrt{\alpha_t}\sqrt{1-\Bar{\alpha}_t}} \epsilon_{\theta}) \label{eq:20x} \\
    % &= (\sqrt{\Bar{\alpha}_{t-1}}\hat{\boldsymbol{x}}_0 + \frac{\sqrt{\alpha_t - \Bar{\alpha}_t}\sqrt{\alpha_t - \Bar{\alpha}_t}}{\sqrt{\alpha_t}\sqrt{1-\Bar{\alpha}_t}} \epsilon_{\theta}) \label{eq:20x} \\
    &= \sqrt{\Bar{\alpha}_{t-1}} \hat{\boldsymbol{x}}_0 + \sqrt{\frac{\alpha_t - \Bar{\alpha}_t}{1- \Bar{\alpha}_t}} \sqrt{1-\Bar{\alpha}_{t-1}} \epsilon_{\theta} \label{eq:21} \\
    &\approx \sqrt{\Bar{\alpha}_{t-1}} \hat{\boldsymbol{x}}_0 + \sqrt{1-\Bar{\alpha}_{t-1}} \epsilon_{\theta} \label{eq:22}
\end{align}
with $\sqrt{\frac{\alpha_t - \Bar{\alpha}_t}{1- \Bar{\alpha}_t}} \approx 1$ for most $t \in (1, T)$.\footnote{Specifically, we adopt a cosine schedule for $\Bar{\alpha}_t$ \citep{nichol2021improved}, and $\sqrt{\frac{\alpha_t - \Bar{\alpha}_t}{1- \Bar{\alpha}_t}} > 0.98$ for 98\% of all $t$, with some outliers as $t \to 0$ and $t \to T$.}

Noting the format simlarity between \autoref{eq:1} and \autoref{eq:22}, we therefore interpret the DDPM decoding transition from $\boldsymbol{x}_t$ to $\boldsymbol{x}_{t-1}$ as (1) predicting an imaginary $\hat{\boldsymbol{x}}_0$, and (2) applying a \emph{compensating} forward diffusion step with a deterministic noise $\epsilon_{\theta}$.

Our decoding strategy in \autoref{eq:18} is in a very similar form as \autoref{eq:22}. We also predict the initial data representation with $\theta$ and apply a forward diffusion step. The difference is that we sample a noise $\boldsymbol{z}$ instead of using the deterministic $\epsilon_{\theta}$, to encourage exploration.

\section{Detailed setup of the comparison with Diffusion-LM \citep{Li2022DiffusionLMIC}}
\label{sec:appendix_comparison_difflm}
We apply block concatenation on ROCStories similarly as OpenWebText, resulting in 50K training sequences of 100 tokens. 
We train Diffusion-LM with a default batch size of 64, learning rate of 1e-4, and 400K steps. We train \methodname{} with a batch size of 512, learning rate of 1e-4, and 20K steps. Both models use a tokenizer of BERT-base-uncased. 
For \methodname{}, additional hyperparameters like decoding block size and one-hot constant remain the same as the main \methodname{} benchmarked with GPT-2. 
For Diffusion-LM, the evaluation in the main paper is an infilling task. We use same decoding hyperparameters as \citet{Li2022DiffusionLMIC}. 
For \methodname{}, the evaluation is a block-wise generation problem with $m$=2 iterations. 
The result of \methodname{} in \autoref{tab:comparison_with_difflm} is obtained with a decoding configuration of $T_{\text{decode}}$=2500 and top-$p$=0.5.

Our \methodname{} in this subsection is initialized with BERT. For a fair comparison, apart from the default Diffusion-LM reported in \autoref{tab:comparison_with_difflm}, we train another Diffusion-LM initialized with the encoder weights of BERT. However, this leads to degenerated results that are much worse than the default Diffusion-LM and our \methodname{}: a MAUVE score of 0.4 out of 100 and a PPL of 73157. This problem is not due to overfitting, as all checkpoints of the model show the same degenerated result. 
Since \citet{Li2022DiffusionLMIC} did not explore this setup in their original work as well, we conjecture that Diffusion-LM may be incompatible with pretrained weights from existing non-diffusion models by nature, a disadvantage to our \methodname{}.

\section{Additional results}
\label{sec:appendix_results}

\begin{figure*}[t]
    \centering
    \begin{subfigure}[h]{0.32\textwidth}
        \includegraphics[width=\textwidth]{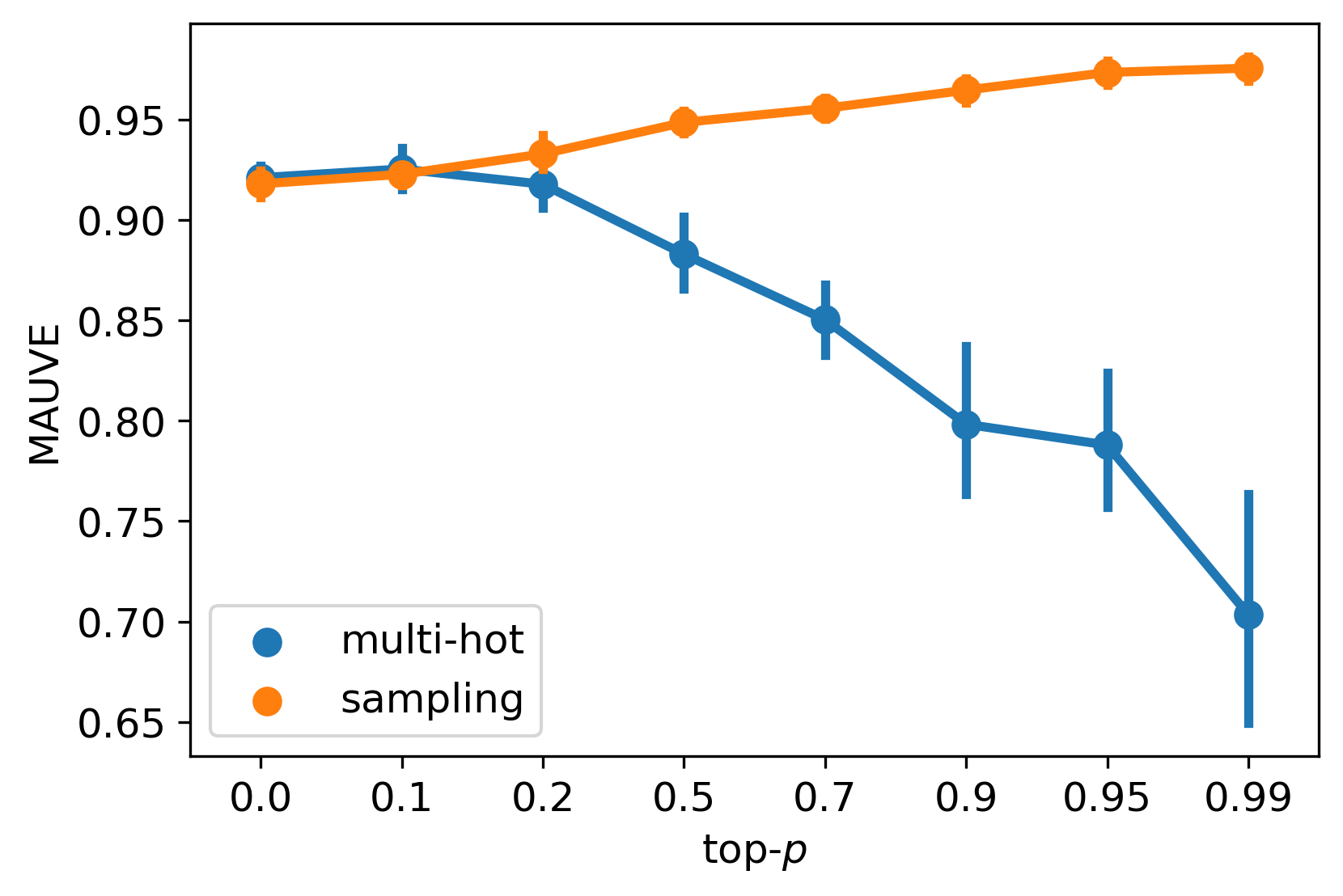}
        % \caption{TBD}
        \label{fig:1-1}
    \end{subfigure}
    \hfill
    \begin{subfigure}[h]{0.32\textwidth}
        \includegraphics[width=\textwidth]{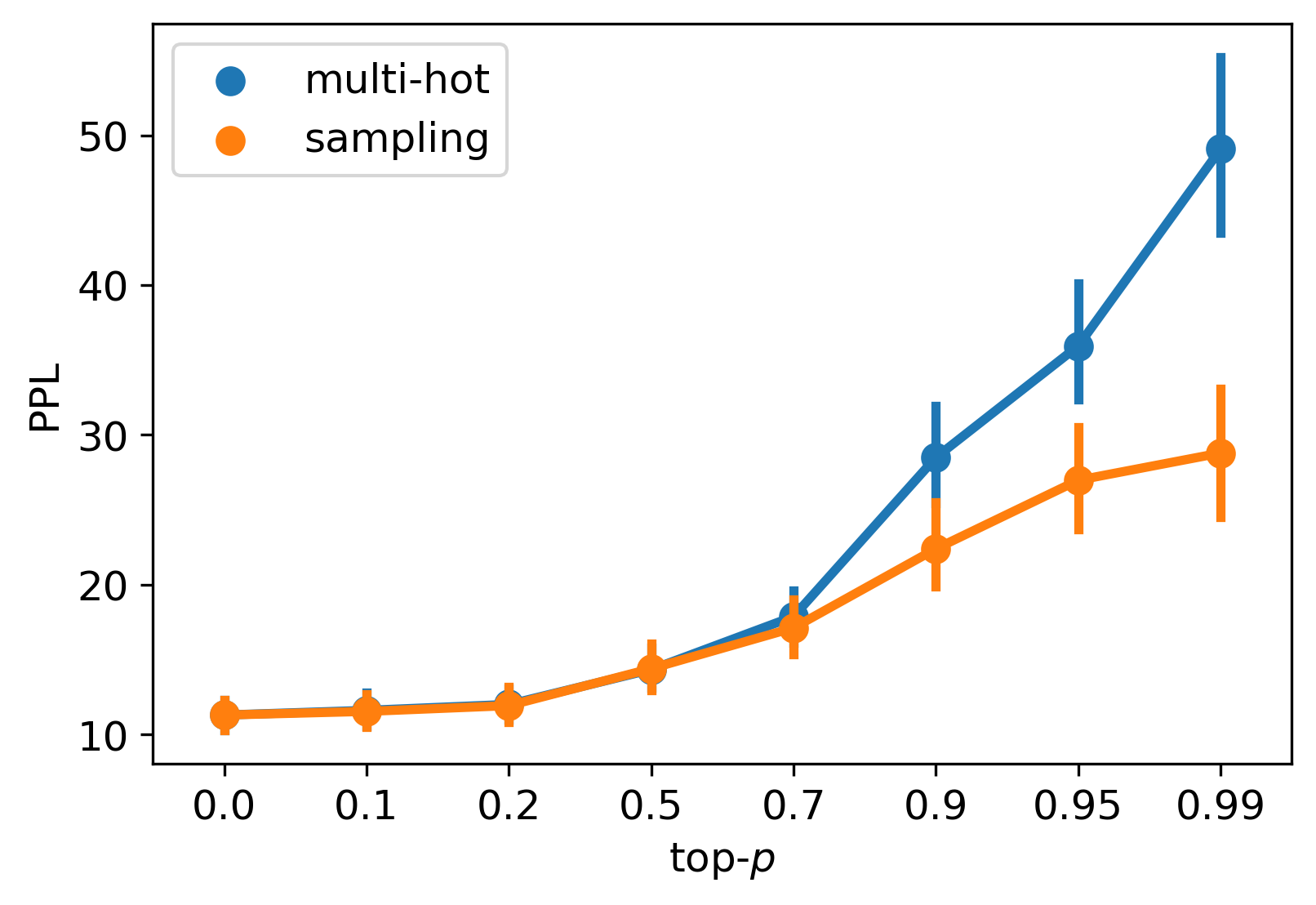}
        % \caption{TBD}
        \label{fig:1-2}
    \end{subfigure}
    \hfill
    \begin{subfigure}[h]{0.32\textwidth}
        \includegraphics[width=\textwidth]{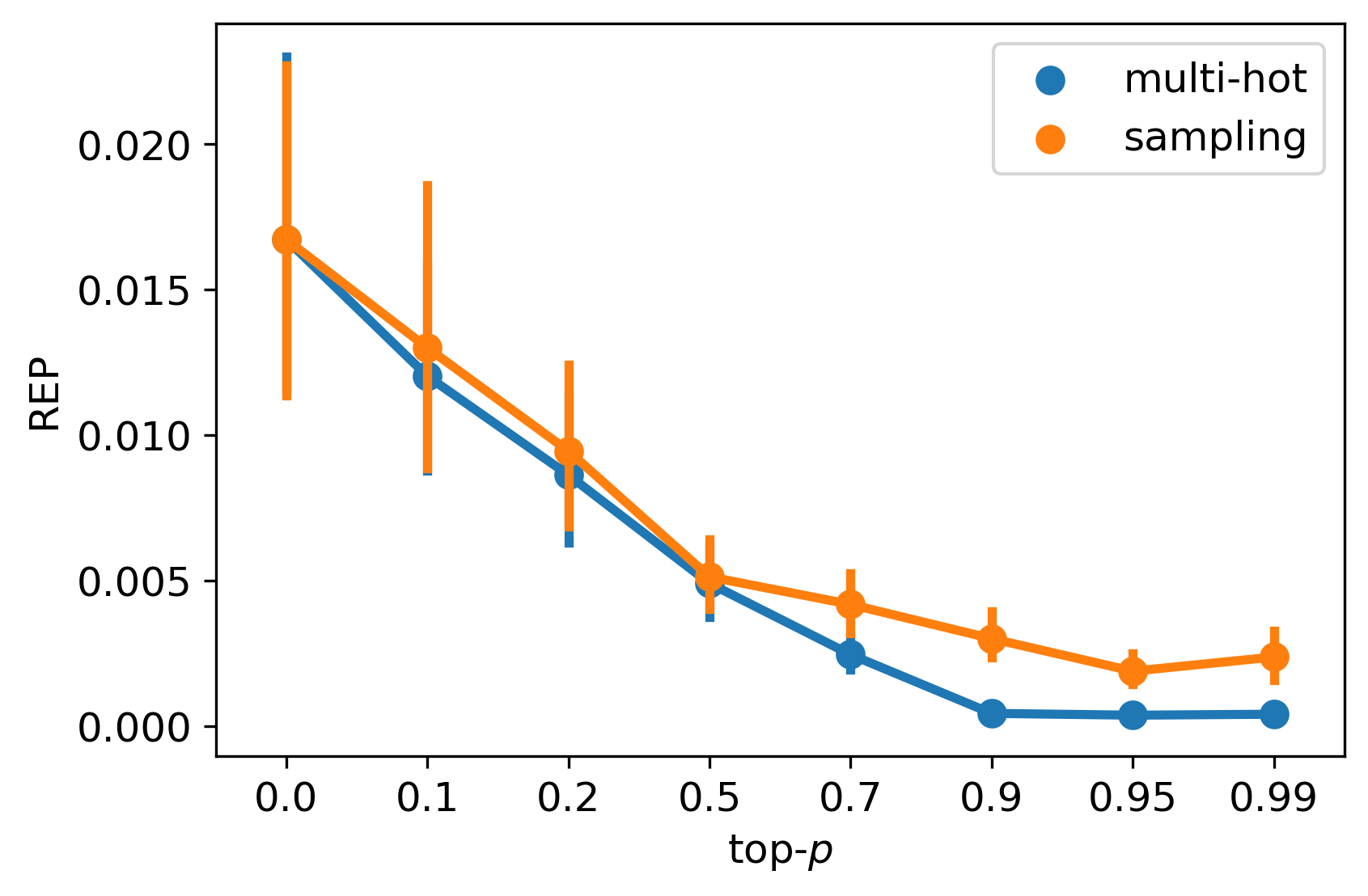}
        % \caption{TBD}
        \label{fig:2-2}
    \end{subfigure}

    % \vfill

    % \begin{subfigure}[h]{0.32\textwidth}
    %     \includegraphics[width=\textwidth]{resources/zipf.png}
    %     % \caption{TBD}
    %     \label{fig:2-1}
    % \end{subfigure}
    % \hfill
    % \begin{subfigure}[h]{0.32\textwidth}
    %     \includegraphics[width=\textwidth]{resources/dist.png}
    %     % \caption{TBD}
    %     \label{fig:1-3}
    % \end{subfigure}
    % \hfill
    % \begin{subfigure}[h]{0.32\textwidth}
    %     \includegraphics[width=\textwidth]{resources/bleu.png}
    %     % \caption{TBD}
    %     \label{fig:2-3}
    % \end{subfigure}
    \caption{Influence of different decoding logits projection strategies and associating top-$p$ for \methodname{} on various text quality metrics. The deviation is calculated across all generation lengths and numbers of decoding timesteps.}
    \label{fig:6_vis}
\end{figure*}

\autoref{fig:6_vis} shows the influence of different logits projection strategies and the associated parameters on the unconstrained generations' output text quality. We observe that reducing top-$p$ $\to$ 0 (greedy projection) can lead to a low perplexity but it is undesirable due to a high repetition rate. We also find the multi-hot projection strategy is overall worse performing than the sampling projection strategy in our setup, indicating it is better to commit the intermediate states to single rather than multiple tokens. 
This can be because our logits mapping involves putting probability mass on singular tokens. The multi-hot projection may still be a viable strategy if future work uses multi-hot logits mapping for the input tokens.

\begin{figure}[h]
    \centering
    \begin{subfigure}[h]{0.36\textwidth}
        \includegraphics[width=\textwidth]{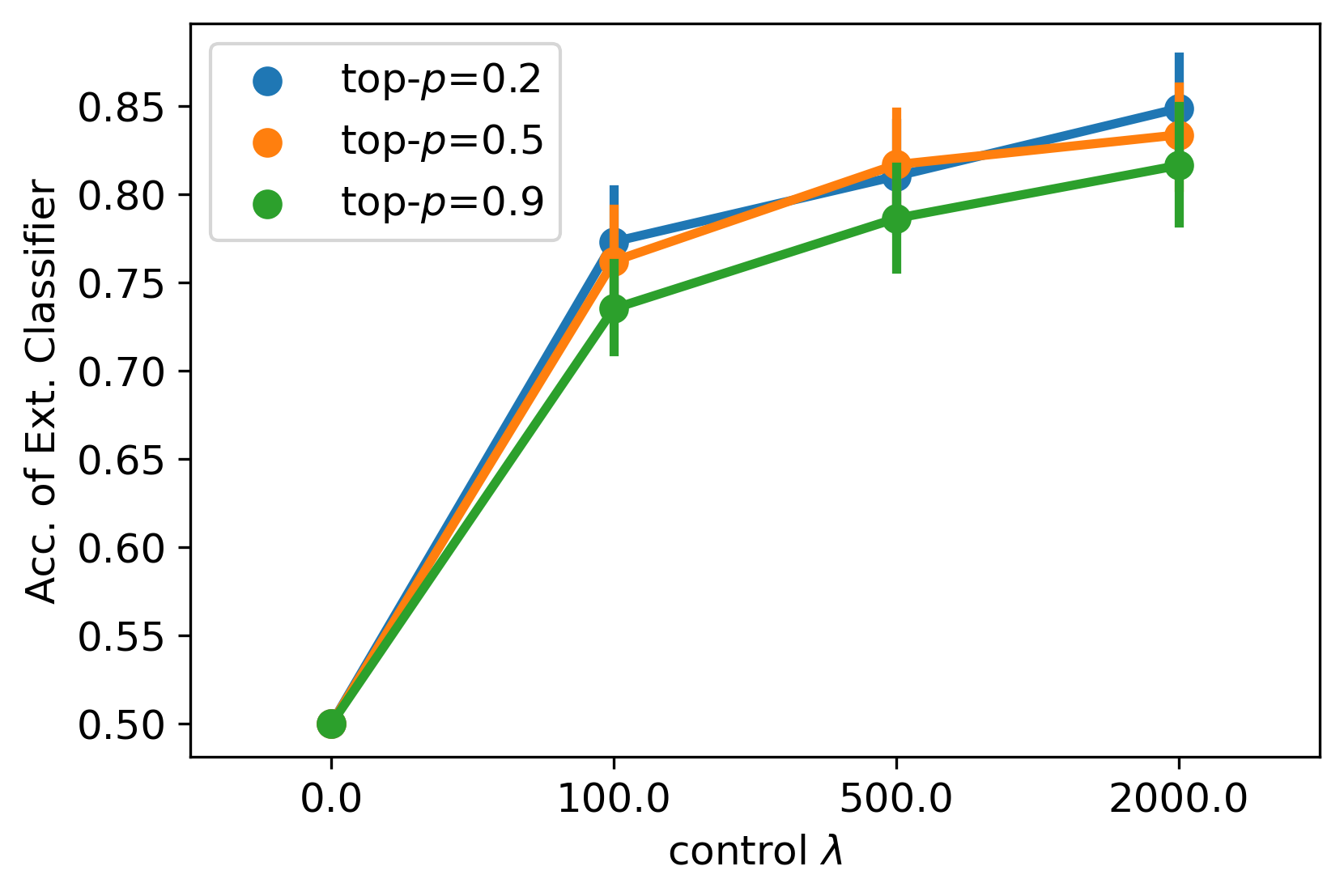}
        % \caption{TBD}
        \label{fig:1}
    \end{subfigure}

    \vfill

    \begin{subfigure}[h]{0.36\textwidth}
        \includegraphics[width=\textwidth]{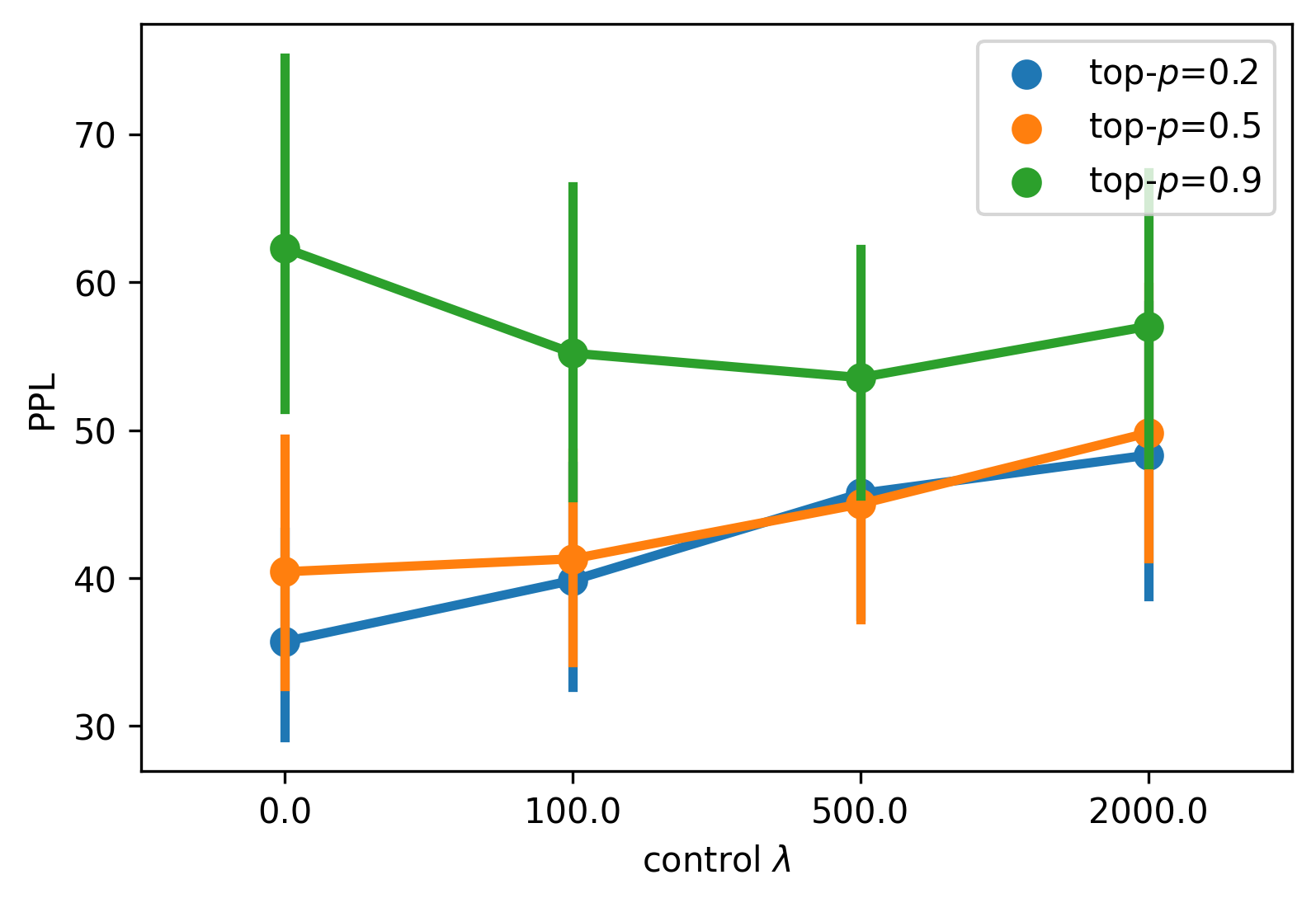}
        % \caption{TBD}
        \label{fig:2}
    \end{subfigure}
    
    \caption{Influence of different control weight $\lambda$ and different top-$p$. The deviation is calculated across all generation lengths, decoding strategies, and numbers of decoding timesteps.
    % \han{updated figure with new PPL}
    }
    \label{fig:2_vis}
\end{figure}

\autoref{fig:2_vis} shows the impact of the control weight $\lambda$ and top-$p$ on the attribute accuracy and perplexity in controlled text generation. As expected, a larger control weight leads to a better external classifier accuracy. The perplexity at the same time increases with a larger $\lambda$, but under a reasonable range for a top-$p$ of 0.2 and 0.5.

\begin{figure}[h]
    \centering
    \includegraphics[width=0.48\textwidth]{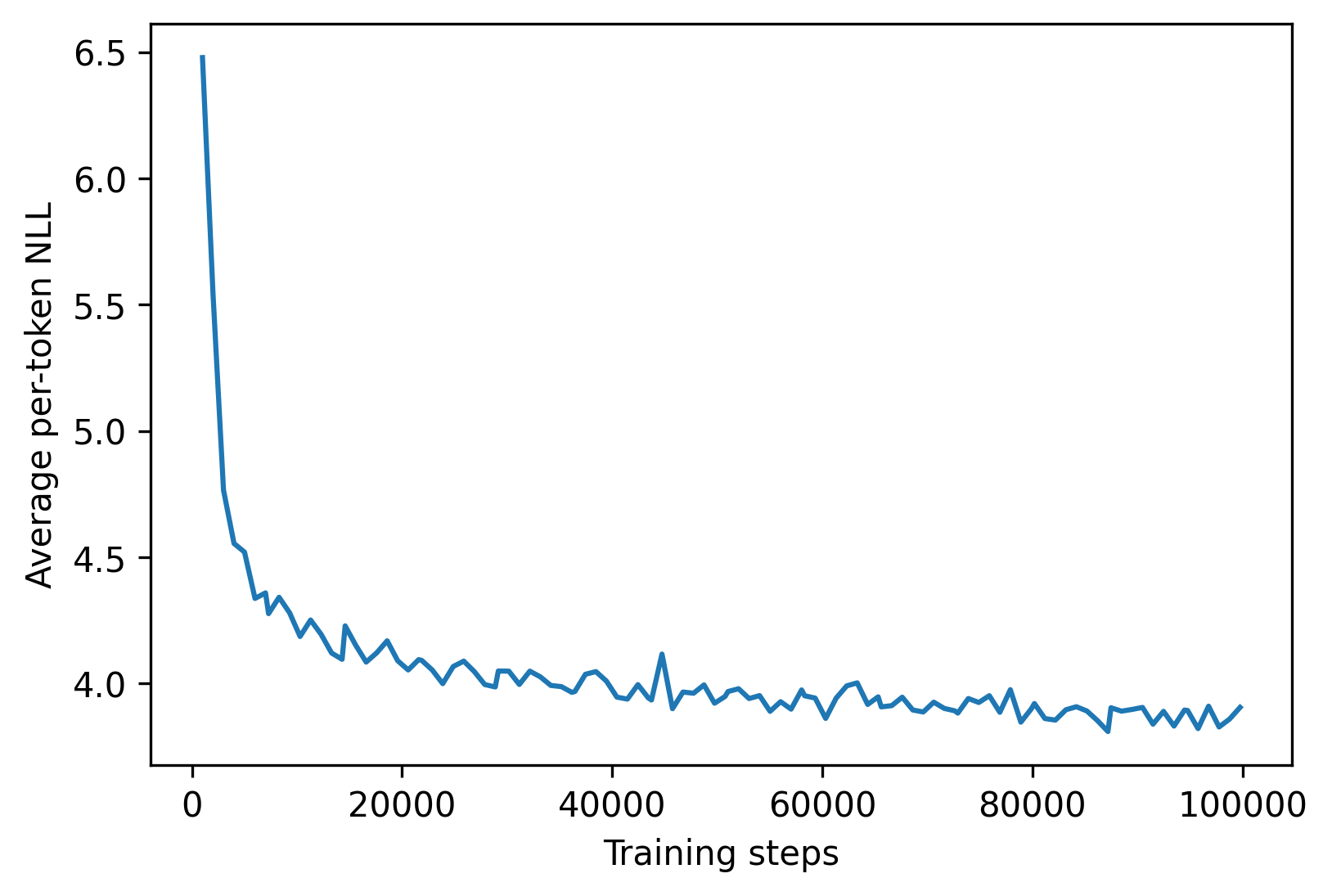}
    \caption{Per-token negative log-likelihood during \methodname{}'s pretraining.
    }
    \label{fig:pt_loss}
\end{figure}

\autoref{fig:pt_loss} shows the pretraining loss trajectory. \autoref{tab:main_result_2}, \autoref{tab:main_result_3}, \autoref{tab:ctr_result_2}, and \autoref{tab:ctr_result} show additional evaluation results of \methodname{} generations. \autoref{tab:qual_ex} and \autoref{tab:traj_qual_ex} show qualitative examples of \methodname{} generations.

\begin{table*}[t]
    \centering
    \begin{tabular}{@{}lp{0.60in}p{0.35in}p{0.42in}p{0.40in}p{0.40in}p{0.40in}p{0.32in}p{0.32in}@{}}
    \toprule
     (Length 25) & MAUVE $\uparrow$ & PPL \small{$\xrightarrow[\text{gold}]{}$} & \small{$|\Delta_{\log \text{PPL}}|$} \normalsize{$\downarrow$} & Dist-1 $\uparrow$ & Dist-2 $\uparrow$ & Dist-3 $\uparrow$ & Zipf \small{$\xrightarrow[\text{gold}]{}$} & Rep $\downarrow$ \\% & BLEU \\
      \midrule
      \textit{Gold continuation} & 100.00 & 21.24  & 0.00 & 93.93 & 93.54 & 88.23  & 0.84 & 0.10 \\% & 100.00 \\
      [6pt]
      \underline{GPT2-medium} \tiny{(Best config)} \\
      \small{Top-$p$=0.95} & 97.35\tiny{$\pm$ 0.29} & 14.31 \qquad \tiny{$\pm$ 0.07} & 0.39 & 73.63 \qquad \tiny{$\pm$ 0.11} & 90.44 \qquad \tiny{$\pm$ 0.13} & \textbf{87.75} \qquad \tiny{$\pm$ 0.13} & 1.01 & 0.21 \qquad \tiny{$\pm$ 0.05} \\% & 2.12 \\
      \underline{GPT2-large} \tiny{(Best config)} \\
      \small{Top-$p$=0.95} & 97.01\tiny{$\pm$ 0.56} & 12.14 \qquad \tiny{$\pm$ 0.06} & 0.55 & 71.94 \qquad \tiny{$\pm$ 0.10} & 89.84 \qquad \tiny{$\pm$ 0.06} & 87.66 \qquad \tiny{$\pm$ 0.06} & 1.02 & 0.23 \qquad \tiny{$\pm$ 0.08} \\% & 2.31 \\
      \underline{GPT2-xl} \tiny{(Best config)} \\
      \small{Top-$p$=0.95} & 97.29\tiny{$\pm$ 0.80} & 11.90 \qquad \tiny{$\pm$ 0.09} & 0.57 & 72.02 \qquad \tiny{$\pm$ 0.04} & 89.58 \qquad \tiny{$\pm$ 0.14} & 87.39 \qquad \tiny{$\pm$ 0.13} & 1.00 & 0.22 \qquad \tiny{$\pm$ 0.02} \\% & \textbf{2.80} \\
      [2pt]
      \underline{\methodname{}-``medium''} \tiny{(Top-3)} \\
      [2pt]
      \small{Sampling $p$=0.99, $T$=1000} & \textbf{98.41} & 38.30 & 0.58 & \textbf{75.61} & \textbf{90.85} & 87.58 & 0.98 & \textbf{0.10} \\% & 2.62 \\
      [2pt]
      \small{Sampling $p$=0.99, $T$=2500} & 98.33 & 30.89 & \textbf{0.37} & 75.04 & 90.64 & 87.54 & 1.02 & 0.18 \\% & \textbf{2.65} \\ 
      [2pt]
      \small{Sampling $p$=0.95, $T$=1000} & 98.18 & 33.79 & 0.46 & 74.70 & 90.67 & 87.62 & 0.99 & 0.18 \\% & 2.53 \\
      \bottomrule
    \end{tabular}
    \caption[Caption for LOF]{Unconstrained generation evaluation of \methodname{} and GPT-2 models at length 25. %Indication of better quality: MAUVE $\uparrow$, PPL $\rightarrow$ PPL$_{\text{gold}}$, Dist-1/2/3 $\uparrow$, Zipf $\rightarrow$ Zipf$_{\text{gold}}$, Rep $\downarrow$, BLEU $\uparrow$. 
    PPL is computed with GPT-Neo-1.3B \citep{gpt-neo}. For GPT-2 models, the results are averaged across 5 random seeds, and we show the best sampling parameter configuration. For our \methodname{}, we show the top-3 configurations. All configurations are ranked based on MAUVE, with original parameters from \citet{Pillutla2021MAUVEMT}.
    }
    \label{tab:main_result_2}
\end{table*}

\begin{table*}[t]
    \centering
    \begin{tabular}{@{}lp{0.60in}p{0.35in}p{0.42in}p{0.40in}p{0.40in}p{0.40in}p{0.32in}p{0.32in}@{}}
    \toprule
     (Length 100) & MAUVE $\uparrow$ & PPL \small{$\xrightarrow[\text{gold}]{}$} & \small{$|\Delta_{\log \text{PPL}}|$} \normalsize{$\downarrow$} & Dist-1 $\uparrow$ & Dist-2 $\uparrow$ & Dist-3 $\uparrow$ & Zipf \small{$\xrightarrow[\text{gold}]{}$} & Rep $\downarrow$ \\% & BLEU \\
      \midrule
      \textit{Gold continuation} & 100.00 & 14.83  & 0.00 & 81.40 & 96.21 & 96.12  & 0.90 & 0.20 \\% & 100.00 \\
      [6pt]
      \underline{GPT2-medium} \tiny{(Best config)} \\
      \small{Top-$p$=0.95} & 97.54\tiny{$\pm$ 0.43} & 11.68 \qquad \tiny{$\pm$ 0.03} & \textbf{0.23} & 58.48 \qquad \tiny{$\pm$ 0.02} & 90.82 \qquad \tiny{$\pm$ 0.04} & 94.56 \qquad \tiny{$\pm$ 0.03} & 1.01 & 0.50 \qquad \tiny{$\pm$ 0.10} \\% & 3.63 \\
      \underline{GPT2-large} \tiny{(Best config)} \\
      \small{Top-$p$=0.95} & 97.36\tiny{$\pm$ 0.22} & 9.43 \qquad \tiny{$\pm$ 0.03} & 0.45 & 56.96 \qquad \tiny{$\pm$ 0.11} & 89.43 \qquad \tiny{$\pm$ 0.10} & 93.96 \qquad \tiny{$\pm$ 0.09} & 1.02 & 0.60 \qquad \tiny{$\pm$ 0.06} \\% & 4.04 \\
      \underline{GPT2-xl} \tiny{(Best config)} \\
      \small{Top-$p$=0.95} & 97.53\tiny{$\pm$ 0.34} & 9.17 \qquad \tiny{$\pm$ 0.04} & 0.48 & 57.10 \qquad \tiny{$\pm$ 0.11} & 89.35 \qquad \tiny{$\pm$ 0.09} & 93.76 \qquad \tiny{$\pm$ 0.08} & 1.00 & 0.58 \qquad \tiny{$\pm$ 0.06} \\% & \textbf{4.40} \\
      [2pt]
      \underline{\methodname{}-``medium''} \tiny{(Top-3)} \\
      [2pt]
      \small{Sampling $p$=0.95, $T$=1000} & \textbf{97.67} & 23.38 & 0.45 & 60.17 & 91.30 & 94.89 & 1.02 & \textbf{0.30} \\% & 3.83 \\
      [2pt]
      \small{Sampling $p$=0.99, $T$=2500} & 97.36 & 21.17 & 0.35 & 60.02 & 90.93 & 94.52 & 1.04 & 0.44 \\% & \textbf{3.92} \\ 
      [2pt]
      \small{Sampling $p$=0.99, $T$=1000} & 97.10 & 26.41 & 0.57 & \textbf{61.26} & \textbf{91.91} & \textbf{95.11} & 1.01 & 0.32 \\% 3.78 \\
      \bottomrule
    \end{tabular}
    \caption[Caption for LOF]{Unconstrained generation evaluation of \methodname{} and GPT-2 models at length 100. %Indication of better quality: MAUVE $\uparrow$, PPL $\rightarrow$ PPL$_{\text{gold}}$, Dist-1/2/3 $\uparrow$, Zipf $\rightarrow$ Zipf$_{\text{gold}}$, Rep $\downarrow$, BLEU $\uparrow$. 
    PPL is computed with GPT-Neo-1.3B \citep{gpt-neo}. For GPT-2 models, the results are averaged across 5 random seeds, and we show the best sampling parameter configuration. For our \methodname{}, we show the top-3 configurations. All configurations are ranked based on MAUVE, with original parameters from \citet{Pillutla2021MAUVEMT}.
    }
    \label{tab:main_result_3}
\end{table*}

\begin{table}[t]
    \centering
    \begin{tabular}{@{}p{0.9in}p{0.58in}p{0.38in}p{0.68in}@{}}
    \toprule
      (Length 12) & C-Ext.\tiny{(Int.)} & PPL & Dist-1/2/3 \\
      \midrule
      \underline{DAPT}$^{\mathbb{CM}}$ & 66.7 & 106.5  & 65/85/79 \\
      [2pt]
      \underline{PPLM}$^{\mathbb{CC}}$ & 58.0 \tiny{(71.7)} & 113.1  & - \\
      [2pt]
      \underline{FUDGE}$^{\mathbb{CC}}$ & 62.6 & \textbf{12.5}  & 52/76/77 \\
      [2pt]
      \underline{GeDi}$^{\mathbb{CM}}$ & \textbf{93.6} & 460.6  & 65/76/69 \\
      [2pt]
      \underline{DExperts}$^{\mathbb{CM}}$ & 87.4 & 69.0  & 65/85/80 \\
      [2pt]
      \underline{MuCoLa}$^{\mathbb{CC}}$ & 89.0 & 38.7  & 49/72/73 \\
      \midrule
      [2pt]
      \underline{M\&M LM$^{\mathbb{HMC}}$} & 65.1 \tiny{(94.3)} & 264.1  & - \\
      [2pt]
      \underline{\methodname{}$^{\mathbb{HMC}}$} & \textbf{79.3} \tiny{(90.5)} & \textbf{58.1}  & 60/83/80 \\ % total ppl 37.2
      \bottomrule
    \end{tabular}
    \caption[Caption for LOF]{Controlled text generation results of \methodname{} and baselines at length 12.  We report the external classifier's accuracy (C-Ext.) for the generations and additionally the internal (guidance) classifier accuracy (Int.) if available. The perplexity (PPL) is computed with GPT2-xl. MuCoLa is the version using two discriminators. 
    $\mathbb{CM}$ stands for customized language model, $\mathbb{CC}$ stands for customized classifier, and $\mathbb{HMC}$ stands for highly-modular classifier (in an order of increasing modularity). Best of $\mathbb{HMC}$ results and all results are bolded. 
    }
    \label{tab:ctr_result_2}
\end{table}

\begin{table}[t]
    \centering
    \begin{tabular}{@{}p{0.9in}p{0.58in}p{0.38in}p{0.68in}@{}}
    \toprule
      (Length 20) & C-Ext.\tiny{(Int.)} & PPL & Dist-1/2/3 \\
      \midrule
      \underline{DAPT}$^{\mathbb{CM}}$ & 70.0 & 78.7  & 64/89/86 \\
      [2pt]
      \underline{PPLM}$^{\mathbb{CC}}$ & 57.6 \tiny{(74.5)} & 61.1  & - \\
      [2pt]
      \underline{FUDGE}$^{\mathbb{CC}}$ & 61.3 & \textbf{10.4}  & 51/80/84 \\
      [2pt]
      \underline{GeDi}$^{\mathbb{CM}}$ & \textbf{96.5} & 190.5  & 70/86/82 \\
      [2pt]
      \underline{DExperts}$^{\mathbb{CM}}$ & 87.1 & 52.3  & 62/89/87 \\
      [2pt]
      \underline{MuCoLa}$^{\mathbb{CC}}$ & 88.3 & 30.3  & 50/76/77 \\
      \midrule
      [2pt]
      \underline{M\&M LM$^{\mathbb{HMC}}$} & 65.9 \tiny{(96.3)} & 167.2  & - \\
      [2pt]
      \underline{\methodname{}$^{\mathbb{HMC}}$} & \textbf{88.0} \tiny{(95.6)} & \textbf{41.6}  & 56/86/87 \\ % total ppl 31.2\\
      \bottomrule
    \end{tabular}
    \caption[Caption for LOF]{Controlled text generation results of \methodname{} and baselines at length 20. We report the external classifier's accuracy (C-Ext.) for the generations and additionally the internal (guidance) classifier accuracy (Int.) if available. The perplexity (PPL) is computed with GPT2-xl. MuCoLa is the version using two discriminators. 
    $\mathbb{CM}$ stands for customized language model, $\mathbb{CC}$ stands for customized classifier, and $\mathbb{HMC}$ stands for highly-modular classifier (in an order of increasing modularity). Best of $\mathbb{HMC}$ results and all results are bolded. 
    }
    \label{tab:ctr_result}
\end{table}

\begin{table*}[ht]
\begin{center}
\renewcommand{\arraystretch}{1.0}
\begin{tabular}{p{0.47\textwidth}p{0.47\textwidth}}
    \toprule
    {Context} & {Generations}\\
    \midrule
    {\begin{lstlisting}
 called the Grand Finale, where it will end its long life by plunging into Saturn's atmosphere this September. Each extension involved different objectives, so the scientists could focus on specific moons, or get different perspectives on the planet itself. This last phase
    \end{lstlisting}} & {\begin{lstlisting}
 of the mission is different altogether, and it is expected to capture the last moments of Cassini orbit. As Wired reports:\n\nThe timing of Saturn's final working flight is not known yet, but Cassini probably has the atmosphere to leave the
    \end{lstlisting}}\\
    {} & {\begin{lstlisting}
 is meant to fill out the entire totality of the mission.\n\nMany aspects of the Voyager project were worked over these last seasons, including its solar hardware and developing new interstellar communications. More is still going to be revealed on the website as you get
    \end{lstlisting}}\\
    \midrule
    {\begin{lstlisting}
 caution.\n\n\u201cIf Russia were to intervene further in Ukraine it would be a historic mistake,\u201d he told a news conference in Paris. \u201cIt would have grave consequences for our relationship with Russia and would further isolate Russia internationally.
    \end{lstlisting}} & {\begin{lstlisting}
\u201d\n\nIn addition to EU sanctions against Russian companies at the ports and other targets of the bloc, Hollande said he was concerned by Russian military involvement in the pro-Russian conflict, which lawmakers said had transformed Ukraine into a new \"post-
    \end{lstlisting}}\\
    {} & {\begin{lstlisting}
\u201d\n\nThe breakthrough has sharpened Moscow\u2019s meddling in pro-Western eastern Ukraine and put Moscow\u2019s relationship with Washington and western Europe on edge after the death of U.S. Col. Chris Stevens.\n\nWestern
    \end{lstlisting}}\\
    \midrule[2pt]
    {\begin{lstlisting}
\n\nThe city
    \end{lstlisting}} & {\begin{lstlisting}[mathescape]
 council will issue a decision in late September on whether to continue efforts to adopt the partnership model at the $\textbf{[NO CONTROL]}$
    \end{lstlisting}}\\
    {} & {\begin{lstlisting}[mathescape]
 is one of the world's fastest-growing cities with over 4 million inhabitants. It is the most $\textbf{[POSITIVE SENTIMENT]}$
    \end{lstlisting}}\\
    {} & {\begin{lstlisting}[mathescape]
 does not have the authority to regulate drug use on public property or punish people for it. The city $\textbf{[NEGATIVE SENTIMENT]}$
    \end{lstlisting}}\\
    \midrule
    {\begin{lstlisting}
\n\nThe movie
    \end{lstlisting}} & {\begin{lstlisting}[mathescape]
\u2019s little-known star, O.J. Simpson, claimed in a lawsuit he had $\textbf{[NO CONTROL]}$
    \end{lstlisting}}\\
    {} & {\begin{lstlisting}[mathescape]
 marks the newest addition to the Marvel Extended Universe and we can't wait to see what's next in $\textbf{[POSITIVE SENTIMENT]}$
    \end{lstlisting}}\\
    {} & {\begin{lstlisting}[mathescape]
 is just another example of the stupid movies that lack an understanding of why writing is important and why it $\textbf{[NEGATIVE SENTIMENT]}$
    \end{lstlisting}}\\
    \bottomrule
\end{tabular} 
\end{center}
\caption{
% This is a placeholder for qualitative examples, for both natural and controlled generation. 
Qualitative examples of \methodname{}'s generations. \emph{Top half}: unconstrained text generation (\Sref{sec:experiments_natural_gen}), given 50 tokens from OpenWebText as the context/prompt and generating the next 50 tokens. We show two prompts and two sample generations for each prompt. \emph{Bottom half}: controlled text generation (\Sref{sec:experiments_controlled_gen}), given prompts from \citet{dathathri2019plug} and generating the next 20 tokens. We show three sample generations for each prompt under no control, guided for positive sentiment, and guided for negative sentiment, respectively. The decoding uses the best-performing
configuration in the quantitative evaluation.
}
\label{tab:qual_ex}
\end{table*}

\begin{table*}[ht]
\begin{center}
\renewcommand{\arraystretch}{1.0}
\begin{tabular}{p{0.05\textwidth}p{0.39\textwidth}p{0.48\textwidth}}
    \toprule
    {$t$} & {$\argmax \boldsymbol{w}_{\text{logits},t}^{c:c+B}$} & {$\argmax \Tilde{\boldsymbol{w}}_{t-1}^{c:c+B}$}\\

    \midrule
    {2500} & {\begin{lstlisting}
 of the to the the the the the the the the the the the the the the the the the the the the the the
    \end{lstlisting}} & {\begin{lstlisting}
apeshifteriao41 fleeting frontman Nutdrop278temp Drama lime Employee cuc rival greatest kan snakes431 cav dreamedRange alloy originally Pact
    \end{lstlisting}}\\
    
    \midrule
    {1500} & {\begin{lstlisting}
 is the to be the, of the,,,\n the the the the the the the the the\n into the.
    \end{lstlisting}} & {\begin{lstlisting}
 stunnedchildrenmetrywaveopensLayer Porn woman transcend242 Homs PluginNext Endsackle microbi spokesperson Brunswick awards":- Sharma Pinball Jr Rug wrapped
    \end{lstlisting}}\\
    
    \midrule
    {1300} & {\begin{lstlisting}
 of the mission it the as a, for the,,, as to as the the moons, and Cass Cassini is
    \end{lstlisting}} & {\begin{lstlisting}
 178 whit promoters du basketballiche SchoolsPur Sack reward basketball corn////WeaponSpeaking squid Chains Caucasian McGivity Me SC rafthr jihadist
    \end{lstlisting}}\\
    
    \midrule
    {1100} & {\begin{lstlisting}
 was based on the in, 2014. Theini will be the the up is the the the the, Hubble but the the
    \end{lstlisting}} & {\begin{lstlisting}
 battles swore starters test thanpadding ambiguityFri BADuitous Stuff depiction bankrupt>>> conversions240Genelvet aptLegweight Riy modesitanesday
    \end{lstlisting}}\\
    
    \midrule
    {900} & {\begin{lstlisting}
 of the Jarminiini Cass Gr, was supposed to be the most ambitious and most attempt to capture all most distant moons
    \end{lstlisting}} & {\begin{lstlisting}
 Sim bag Ves serotonin._ Fab gameplay ransom Alisonorks Fargo expand Rhode pursuing most plagued formulateheter plainly troubled Professional Binary Creek geared
    \end{lstlisting}}\\
    
    \midrule
    {800} & {\begin{lstlisting}
 is all about Saturn. The Eini will, the closest that the instruments have reached will be to stop in on the Saturn
    \end{lstlisting}} & {\begin{lstlisting}
omial allcounter Saturn. The Directthank Ecuador two thelearning that the Animation have brothers will make toousands downtown governance the Further
    \end{lstlisting}}\\
    
    \midrule
    {700} & {\begin{lstlisting}
 will allow the Cass to finally see the planet's relatively small atmosphere and finally be able to procure an accurate way of understanding how
    \end{lstlisting}} & {\begin{lstlisting}
 willPocket prelim Klux to finally see the planet intelligent relatively jumper atmosphere and halted Fly activityvirt00000 trem accurate way of Inferno what
    \end{lstlisting}}\\
    
    \midrule
    {600} & {\begin{lstlisting}
 will allow the scientists to better study the effects of Grand Impact, and also be able to get much more data and images of
    \end{lstlisting}} & {\begin{lstlisting}
 will allowert scientists Damien better study the effects of Grand Impact, andasket bebery to get much more data and images of
    \end{lstlisting}}\\
    
    \midrule
    {500} & {\begin{lstlisting}
 will allow the scientists to better see the interior of its atmosphere, and also be able to get much more knowledge and understanding of
    \end{lstlisting}} & {\begin{lstlisting}
 will allow the scientists to better see the interior of its atmosphere, and also be able to get much more knowledge and understanding of
    \end{lstlisting}}\\
    
    \midrule
    {1} & {\begin{lstlisting}
 will allow the scientists to better see the interior of its atmosphere, and also be able to get much more knowledge and observations of
    \end{lstlisting}} & {\begin{lstlisting}
 will allow the scientists to better see the interior of its atmosphere, and also be able to get much more knowledge and observations of
    \end{lstlisting}}\\

    \bottomrule
\end{tabular} 
\end{center}
\caption{
The intermediate states of generation as $t$ decreases ($T$=2500, $B$=25, top-$p$-sampling=0.99). The context $\boldsymbol{w}^{<c}$ here is the first example prompt in \autoref{tab:qual_ex}: ``{\small{\texttt{ called the Grand Finale, where it will end its long life by plunging into Saturn's atmosphere this September. Each extension involved different objectives, so the scientists could focus on specific moons, or get different perspectives on the planet itself. This last phase}}}''. 
There is no change in the outputs during $500 > t > 1$. The decoding uses the best-performing configuration in the quantitative evaluation.
}
\label{tab:traj_qual_ex}
\end{table*}

\end{document}